\documentclass[pdflatex,hidelinks]{sn-jnl}
\usepackage{doi}

\usepackage{graphicx}%
\usepackage{multirow}%
\usepackage{amsmath,amssymb,amsfonts}%
\usepackage{amsthm}%
\usepackage{mathrsfs}%
\usepackage[title]{appendix}%
\usepackage{xcolor}%
\usepackage{textcomp}%
\usepackage{manyfoot}%
\usepackage{booktabs}%
\usepackage{algorithm}%
\usepackage{algorithmicx}%
\usepackage{algpseudocode}%
\usepackage{listings}%
\usepackage[authoryear, sectionbib]{natbib}
\usepackage{bibunits}
\usepackage{rotating}
\usepackage{subcaption}
\defaultbibliographystyle{apsr}
\usepackage{tabularx} 

\usepackage{setspace}

\theoremstyle{thmstyleone}%
\theoremstyle{thmstyletwo}%
\usepackage{tikz}
\usetikzlibrary{arrows.meta,positioning,fit,calc,shapes.misc,patterns}

\theoremstyle{thmstylethree}%
\begin{document}

\title {Survey Transfer Learning: Recycling Data with Silicon Responses}

\author[]{\fnm{Ali} \sur{Amini}}\email{aa6718a@american.edu}

\affil[] {\orgdiv{Department of Government}, \orgname{American University}, \orgaddress{\street{4400 Massachusetts Avenue, NW}, \city{Washington}, \postcode{20016}, \state{DC}, \country{United States}}}

\abstract{
As researchers increasingly turn to large language models (LLMs) to generate synthetic survey data, less attention has been paid to alternative AI paradigms given environmental costs of LLMs. This paper introduces Survey Transfer Learning (STL), which develops transfer learning paradigms from computer science for survey research to recycle existing survey data and generate empirically grounded ``silicon responses.” Inspired by political behavior theory, STL leverages shared demographic variables with high predictive power in a polarized American context to transfer knowledge across surveys. Using a neural network pre-trained on the Cooperative Election Study (CES) 2020, freezing early layers to preserve learned structure, and fine-tuning top layers on the American National Election Studies (ANES) 2020, STL generates ``silicon responses”  CES 2022 and in held-out ANES 2020 data with accuracy rates of up to 93 percent. Results show that STL outperforms LLMs, especially on sensitive measures such as racial resentment. While LLMs’ silicon samples are costly and opaque, STL generates empirically grounded silicon responses with high individual-level accuracy, potentially helping to mitigate key challenges in social science and the polling industry.}

\maketitle

\section*{Introduction}\label{sec1}

At a time of expanding computational power, intensifying environmental concerns about emerging generative AI \citep{Ren2024LLMEnvironment, zewe2025generative}, and mounting challenges to traditional survey research \citep{Bailey2024PollingAtACrossroads}, political analysts can and must make better use of existing data by recycling survey data directly with deep learning algorithms and specifically transfer learning paradigms. Large language models, while powerful for text-based applications, are not well-suited for structured data analysis. Structured data such as survey data consists of predefined schemas and organized formats, which makes data more manageable and facilitates analysis and processing by machines \citep{chen2020hybridqa, sui2024table}. American public opinion data is a particularly strong candidate for directly applying deep learning algorithms instead of using large language models, as the rich survey responses are inherently structured, with defined variables, scales, and categorical responses that lend themselves to specialized analytical approaches rather than text generation. This paper demonstrates that recycling American public opinion survey data with targeted deep learning algorithms and specifically transfer learning between surveys offers a more cost-effective, more accurate, and empirically grounded approach to predicting survey responses compared to LLMs, and is more sustainable and environmentally friendly as well.

This sustainability concerns compounds longstanding methodological problems in survey research. Missing data biases inference \citep{King2001Multiple}, while rising costs and declining response rates \citep{Shapiro2019} have led some scholars to experiment with LLM-based ``silicon samples'' \citep{argyle2023out, Broska2025, Sarstedt2024, Demszky2023}. Yet emerging scholarship argues that LLM-generated responses are prone to systematic bias, instability, and a lack of transparency \citep{agnew2024illusion, Rossi2024, bisbee2024synthetic}.

Beyond these data collection challenges, researchers face equally problematic issues in data utilization: the mismatch between available survey questions and the specific variables required for hypothesis testing. Even high-quality, nationally representative surveys often lack key variables of interest, especially interaction terms or additional controls, needed for nuanced regression models.These gaps particularly burden graduate students and early-career scholars, who lack resources to commission surveys or pay for comprehensive question sets. 

Scholars therefore face suboptimal choices: abandon theoretically important tests, substitute proxy variables prone to measurement error, or commission costly new surveys via private firms. These approaches are not only financially and logistically burdensome but also environmentally costly, methodologically less transparent, and lead to respondent fatigue.

To address these challenges, and in light of the environmental costs of generative AI \citep{zewe2025generative}, I introduce Survey Transfer Learning (STL), a sustainable framework for recycling survey data. Adapted from transfer learning in computer science \citep{Shaha2018}, STL applies this paradigm to survey research by leveraging Anchor Transfer Variables (ATVs)—shared demographic and partisan indicators such as age, education, gender, income, race, region, and party identification—to transfer knowledge across nationally representative datasets. In doing so, STL generates empirically grounded ``silicon responses” for missing or unasked items, offering a transparent and statistically valid alternative to both LLMs and repeated survey commissioning.

This paper makes several key contributions: First, it develops transfer learning for survey data for the first time via STL for generating \textit{silicon responses}. Second, it shows that in the current context of sorted American politics \footnote{As detailed in the supplemental materials 1, STL is motivated by theoretical insights from American political behavior theories: a small set of predictors—party identification, ideology, education, race, religion, gender, age, and income, effectively predicts many policy preferences and political behaviors, especially in polarized and sorted electorate.}, ATVs predict effectively in deep learning and STL, and can be applied for methodological adjustments in survey firms and in large-\(N\) surveys such as the ANES. Third, it documents a proof of concept, using real-world data from CES and ANES, that STL outperforms both multiple imputation, intra-survey deep learning and LLM-generated synthetic data in predictive accuracy and stability. Fourth, it calls for more climate-conscious political methodologies by treating surveys as interconnected, reusable data resources rather than isolated instruments with a practical solution. While beyond the scope of this paper, STL has potential for improving proxy finding by data integration between surveys:   As noted by \cite{Rapeli2022, Hobbs2010, Kreuter2010}, proxy finding is a common issue in social science research.

\subsection*{Why American Politics is Ripe for Transfer Learning and Recycling Survey Data?}
Since \textit{The People's Choice}, \cite{lazarsfeld_berelson_gaudet_1944}'s seminal work in 1940,  surveys have been foundational to American political research. Whether examining voting behavior through the lens of Party ID, as in the \textit{Michigan} model \citep{campbell_converse_miller_stokes_1960}, or through sociological contexts, as in the \textit{Columbia} model \citep{berelson_lazarsfeld_mcphee_1954}, nationally representative surveys remain crucial and constant for political behavior scholarship. Despite their foundational role, the enduring potential of large-N surveys remains underutilized, as scholars and journals increasingly prioritize original data collection. At the same time, the survey industry faces mounting methodological and practical challenges—pressing issues that demand innovative solutions grounded in computational advances and the shared structure of existing datasets.

The United States, as the world's wealthiest country and oldest continuous democracy, has federally funded numerous large-scale academic surveys that serve as the foundation of American political behavior research. Notable examples include the American National Election Studies (ANES), run by the University of Michigan and Stanford University \citep{anes_stanford}, and the Cooperative Election Study (CES), managed by Harvard University's Institute for Quantitative Social Science \citep{walsh2024tracking}. Both studies employ rigorous methodological standards and provide invaluable data for exploring ordinary citizens' opinions and testing key political theories \citep{Burns2006History}. These surveys always share common demographic and ideological variables (shared domains) while differing in policy-specific questions (distinct domains), creating ideal conditions for transfer learning (TL). By reusing pre-trained models, TL offers an opportunity to systematically infer responses to unasked questions, mitigating survey crises.

These days, political science faces a multi-faceted crisis that affects the field from multiple perspectives that are beyond the scope of this paper. Just to name a few: From a methodological standpoint, declining response rates and survey fatigue have reduced data quality, while rising costs force researchers to rely on smaller, less representative samples, compromising methodological rigor \citep{meterko2015response, keeter2017lowresponse}. The proliferation of independent scholarly surveys has also led to data fragmentation, making it difficult to synthesize findings across studies. 

Furthermore, online survey panels can introduce biases, such as professional \citep{hillygus2014professional} and self-selected respondents \citep{matthijsse2015internet} who frequently participate in multiple surveys—raising concerns about data quality. From a polling perspective, many polling efforts did not work as intended. Pre-election polls generally failed to accurately predict the level of support for Republican candidates in the 2020 elections \citep{clinton2022reluctant}. The 2020 polls featured an unusual magnitude of polling error, marking the highest level in 40 years for the national popular vote \citep{clinton2021aapor}. Needless to say, the popular vote for Trump in 2020 shocked pollsters and political scientists alike, and similar surprises have emerged in 2024 \citep{clinton2024polls}. This highlights another emerging challenge for public opinion research. These interconnected challenges underscore the potential transformative value of AI-assisted survey methodologies in addressing traditional research limitations.

Beyond methodology, accessibility and inclusivity remain major challenges. Academic publishing often prioritizes original data collection, which disadvantages students and researchers without the financial resources to conduct their own surveys. This journal bias toward original data reinforces inequities in academic opportunities, limiting contributions from scholars at underfunded institutions.

Meanwhile, I argue that sustainability concerns add another layer to this crisis. Running new large-N surveys requires significant computational and digital resources, contributing to digital carbon emissions through server energy consumption, electronic device usage, and the cumulative environmental footprint of large-scale data collection \citep{skare2024digitalization}: An average data-driven company with 100 full-time staff members is likely to produce around 2,203 tons of CO2 emissions annually due to new data \citep{lboro2023co2tool}. 
These issues persist despite our unprecedented access to computational power, highlighting the inefficiencies in how survey data is utilized.

Transfer learning offers a solution by allowing scholars to recycle survey data, extracting new insights from existing datasets rather than continuously running costly and resource-intensive surveys. At the same time, demographic and ideological sorting in American politics has made TL particularly viable. Research on polarization and partisan alignment suggests that a small set of demographic predictors—such as party ID, race, education, age, and ideology—can reliably predict policy preferences \citep{bishop_2008, levendusky_2009}. This growing predictive power allows pre-trained models on CES data to systematically generate responses for missing questions in ANES (and vice versa) with high accuracy.

\subsection*{Transfer Learning between Surveys in Polarized Times} \label{sec3}

Political behavior models have been profoundly shaped by survey research, particularly in demonstrating the explanatory power of demographic variables \citep{bartels_2000, levendusky_2009}. Moreover, research on 'sorting' further underscores the significance of these demographic predictors. \cite{bishop_2008} demonstrates how Americans increasingly cluster into politically homogeneous communities, a trend that accelerates partisan alignment: Democrats and Republicans systematically relocate to areas consistent with their political views \citep{gimpel_hui_2015, sussell_2013}. This ideological sorting is evident in both symbolic ideology (self-identified liberal-conservative placement) and operational ideology (specific policy positions), which have become increasingly correlated with partisan identity \citep{fiorina_abrams_pope_2011, hetherington_2009, weber_klar_2019}.

Polarization, in its various forms, also has become integral to understanding American politics \citep{Abramowitz2008, Benson2024, Iyengar2019, hetherington_2009}. While polarization and sorting pose challenges for democratic theory by intensifying partisan divides, they also create unique AI opportunities for knowledge transfer between datasets. The increased alignment of political attitudes and behaviors with demographic variables enhances their predictive power, making demographic features more robust anchors for imputing missing policy preferences through transfer learning. 

This combination of polarization and sorting presents unique opportunities for leveraging knowledge transfer between datasets. The increased alignment of political attitudes and behaviors with demographic variables enhances their predictive power, making demographic features robust anchors for imputing missing policy preferences through transfer learning. It is important to note that this paper does not merely advocate for a specific application of TL in ANES and CES data; it establishes a new research agenda. I argue that scholars can and must see survey data with common domains (shared variables) but related tasks (prediction) as opportunities for transfer learning.

\section*{From the Challenges of Silicon Samples to an Empirical Solution: Silicon Responses}

The above mentioned survey challenges\footnote{At AAPOR 2025, several panels focused on these issues, noting that response rates in the \textit{New York Times} survey had dropped to 1\% \citep{aapor2025}.} are happening at the peak of artificial intelligence power. The power of AI has created a wave of methodological papers LLMs to address survey issues. In January 2024, for example, the president of the Society for Political Methodology wrote to the listserv noting, ``We are increasingly receiving papers at \textbf{Political Analysis} that engage with \textbf{AI and GPT}, an emerging area of \textit{excitement} for our field,''  seeking reviewers for an unprecedented surge in AI and GPT-related submissions \citep{polmeth_mailing_list}. This \textit{excitement} has centered on using LLMs \citep{aher2023using,argyle2023out} like GPT that could  ``supplant human participants for data collection'' \citep{grossmann2023ai}. Many scholars in public opinion have followed this enthusiasm to generate LLM-synthetic data \citep{Li2023, Veselovsky2023, Josifoski2023}.

However, while still an active and useful research agenda, this enthusiasm after one year has encountered significant pushback. Scholarly discourse has introduced critical caveats and critiques \citep{Pilati2024}, demonstrating that ``substitution proposals'' for using LLMs in place of human participants ignore and ultimately conflict with foundational values in survey research: representation, inclusion, and understanding \citep{agnew2024illusion}. \citet{Rossi2024} review shows that LLM-generated data papers are largely ``motivated by the problem of scarcity'' and  details the epistemological and methodological challenges of LLM-generated data, particularly their inability to capture the nuanced, context-specific nature of human behavior. LLMs can produce outputs that lack intersectional fidelity, marginalizing smaller or vulnerable groups and failing to replicate the diversity of real-world populations \citep{Rossi2024} alongside sexism and racism \citep{Fieck2025BiasInLLMs, Simpson2025ParityBenchmark}\footnote{Beyond significant ethical and legal concerns \citep{hao2024synthetic}, LLMs often amplify biases present in their training data, reinforcing dominant norms while marginalizing minority perspectives. This can perpetuate unfair discrimination and cause both representational and material harm through stereotyping and harmful identity associations \citep{pokotylo2024ethical, weidinger2022taxonomy}.} and a tendency to ``hallucinate'' \citep{mckenna-etal-2023-sources}.

These limitations are compounded by the instability of LLM outputs, which can vary significantly even under identical conditions \citep{Rossi2024}, confirming that ``synthetic data fail even the most basic requirements for replication'' \citep{bisbee2024synthetic}. This instability makes LLM-generated data less reliable for students of political science, particularly given the emphasis on the principle that ``science is a public enterprise'' \citep{king2021designing} : a cornerstone of recent trends toward transparency and the requirement for replicable data \citep{tripp_dion_2024}.

Alongside these methodological and ethical concerns are environmental ones. Training and deploying large-scale AI systems require substantial electricity and water resources. Estimates suggest that data centers supporting global AI infrastructure already consume more water than some countries, and could require several times Denmark’s annual water use in the near future. A single ChatGPT request consumes far more electricity than a typical web search, and in Ireland the energy used by data centers may soon represent more than one-third of national demand \citep{UNEP2024AIEnvironment}. MIT reports that generative AI workloads consume up to seven or eight times more energy than conventional computing, while inference continues to draw growing shares of both power and water \citep{zewe2025generative}. These issues should not deter scholars from working on LLMs, but rather motivate them to recognize these impacts and actively explore alternative areas of artificial intelligence research that pose fewer environmental challenges.\footnote{Instead of creating synthetic personas, this data-driven methodology uses transfer learning to leverage large, representative datasets like ANES, CES, and Pew surveys. Beyond empirical validity, this approach also addresses pressing increasing environmental concerns. Large-scale AI systems rely on data centers whose ecological footprint is staggering: global AI infrastructure already consumes more water annually than some countries, with one estimate projecting it could use six times more water than Denmark in the near future. Training and deploying LLMs require enormous energy inputs—a single ChatGPT request consumes roughly ten times the electricity of a Google Search—contributing to greenhouse gas emissions that in Ireland alone may soon account for 35\% of national energy use. They also generate electronic waste, since producing a 2 kg computer requires 800 kg of raw materials, including rare earth elements mined in ecologically destructive ways \citep{UNEP2024AIEnvironment}.}

Survey Transfer Learning (STL)  responds to these challenges by building directly on existing, representative survey data rather than generating synthetic personas. It reuses information already present in high-quality surveys and transfers knowledge across datasets with shared demographic and partisan predictors. This approach avoids the instability and opacity of LLM-synthetic data, while also reducing the environmental costs of large-scale generative models. STL is therefore offered not as a wholesale replacement for surveys, but as a practical framework for reusing and extending them in ways that are empirically grounded, interpretable, and more sustainable.


\section*{Survey Transfer Learning Framework}

Transfer learning in computer science literature involves pre-training a  model on a large (source) dataset and fine-tuning it on a smaller (target) dataset to optimize performance \citep{taylor2009transfer, pan2010survey, Zhao2024}. While conventional transfer learning application has been applied in other areas of social science, such as computer vision \citep{Torres2022}, annotating data in large language models (LLMs) \citep{Laurer2024}, and detecting deep-fake tweets \citep{Khalid2024}, and hate speech \citep{Yuan2025}, it has not yet been utilized for survey data. This makes this study the first to explore the potential of developing transfer learning in bridging data gaps across political science surveys, offering a novel approach to addressing long-standing data fragmentation challenges.\footnote{One of the innovations of this paper is the argument that we can, and should, treat survey data more like image data in transfer learning: as structured, high-dimensional inputs that are well-suited for transfer learning. Just as images became the foundational input for deep learning breakthroughs, survey data can similarly be transformed through STL into a reusable, flexible input for social science research. This idea, which is welcomed by the APPOR community, can also utilize the rich data from polls.}
    
As shown in Figure 1, the core idea of STL is simple: STL helps  researchers to generate \textit{silicon responses} for unasked policy questions in the target dataset by learning patterns based on ATVs from the source dataset via deep learning architecture. ATVs act as a bridge between surveys, a process computer scientists refer to as ``data integration'' \citep{10.1145/543613.543644}.

\begin{figure}
    \centering
    \includegraphics[width=1\linewidth]{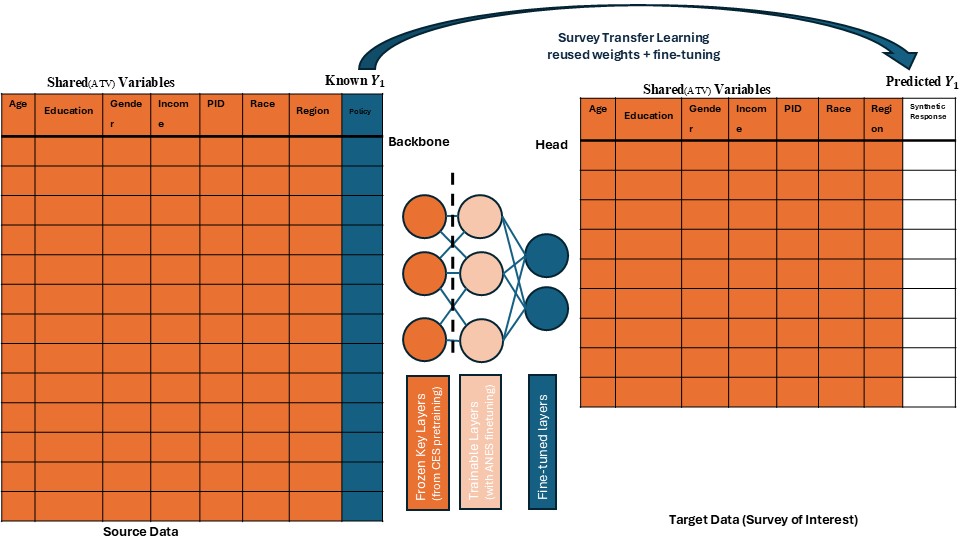}
    \caption{\textbf{Survey Transfer Learning (STL) Framework for Cross-Survey Domain Adaptation.} 
    The STL approach reuses knowledge from a source survey (CES 2020) to predict the same outcome in a target survey (ANES 2020). 
    Both datasets share Anchor Transfer Variables (ATVs)—demographic and ideological features such as age, education, gender, income, party identification (PID), race, and region. 
    In Stage 1, the backbone and head are trained on the source survey to predict a known policy outcome ($Y_1$). 
    In Stage 2, the backbone’s earlier layers (orange) are frozen to preserve CES-learned structure, while later backbone layers (peach) and the task-specific head (blue) are fine-tuned on the target survey. 
    The fine-tuned model then generates synthetic responses ($\hat{Y}_1$) for the target dataset, enabling high-accuracy predictions without re-collecting all outcome data. 
    This design ensures that \textbf{knowledge transfer occurs in the backbone}, while the head remains flexible and task-specific.}
    \label{fig:STL}
\end{figure}

STL offers broad applications beyond this study, including missing data imputation\footnote{See Supplemental Materials 2}, proxy variable generation, and adaptive questionnaire design. This approach can help proxy finding in social science research, enabling researchers to ``borrow" variables from other surveys and create synthetic measures grounded in empirical reality.

These contributions respond to a critical moment in the discipline's evolution. At a time when computational tools are more accessible than ever, and the survey industry faces a crisis without clear consensus on good survey practices, we can begin to reconceptualize surveys, not as standalone instruments, but as dynamic, interconnected sources of insight that can be systematically reused to promote efficiency, inclusivity, and sustainability in research. While scholars currently run individual surveys and fine-tune GPTs with financial resources as the primary constraint, this approach inadvertently creates environmental costs and tends to favor well-funded researchers and institutions, highlighting the need for more inclusive and sustainable research practices.

\section*{Methodology}

Theoretically, transfer learning operates by transferring knowledge between a source domain–task pair \((D_S, T_S)\) and a target domain–task pair \((D_T, T_T)\). In this study, the CES dataset serves as the source domain \((D_S)\) with tasks \((T_S)\) such as predicting outcomes like vote choice. The ANES dataset represents the target domain \((D_T)\), where the task \((T_T)\) involves imputing missing variables based on shared demographic and ideological features. By pre-training models on CES, the framework captures robust relationships within the source domain, which can then be generalized and fine-tuned for the target domain \citep{Goodfellow2016}.

The pre-training phase utilizes CES data to build predictive models that learn feature representations generalizable across related tasks. Fine-tuning adapts these pre-trained models to ANES by leveraging shared features like race and party identification while retaining general patterns from CES. By freezing early model layers and updating later ones, this step allows the model to adjust to ANES-specific nuances while preserving CES-informed knowledge, as practiced in transfer learning applications \citep{yosinski2014transferable}. The shared features between CES and ANES facilitate knowledge transfer, adhering to established transfer learning principles \citep{pan2010survey}.

Unlike synthetic data generation methods, this approach relies on empirical relationships observed in CES to infer lacking variables in ANES. This study’s methodology is inspired by seminal studies on transfer learning. \citet{pan2010survey} provides a comprehensive framework emphasizing cross-domain applications, while \citet{yosinski2014transferable} highlights the adaptability of features learned in one domain for tasks in another. \citet{Goodfellow2016} further discusses the role of pre-training and fine-tuning in optimizing transfer learning workflows. These foundational principles inspire the application of transfer learning to address challenges in survey-based political research.

It is worth noting that I extend the classical transfer learning in computer science by introducing a novel multi-head architecture tailored for political science applications. As illustrated in Figure 1, my pipeline begins with a base model trained on CES data (Task 1), producing predictions for outcome \( Y_1 \) via Head 1. This model is then fine-tuned on ANES data (Task 2) to adapt to a different outcome variable \( Y_2 \). Finally, the same backbone model, with a third specialized head, is applied to a new dataset with shared features to generate synthetic policy responses.

Moreover, these models can be made publicly accessible, allowing researchers to directly apply them for specific tasks, for example, predicting support for Trump in 2020 or measuring racial resentment. There are two key assumptions for effective implementation in the context of American Politics: First, the input dataset should ideally be nationally representative, and second, the target year should align with the model’s training context, as political attitudes in the U.S. evolve significantly over time.

\subsection*{Mathematical Formulation}

I begin with a standard feedforward neural network of depth \(L\):
\[
f(x) : \mathbb{R}^d \to \mathbb{R}, \quad 
f(x) = W_L \,\sigma \!\Big( W_{L-1}\,\sigma \!\big( \cdots \sigma(W_1 x + b_1) \cdots \big) + b_{L-1} \Big) + b_L ,
\]
where \(x \in \mathbb{R}^d\) is the vector of demographic and ideological inputs 
(e.g., race, education, gender, income, party ID, region), 
\(W_\ell, b_\ell\) are the weights and biases at layer \(\ell\), 
and \(\sigma(\cdot)\) is a nonlinear activation (ReLU).

For binary outcomes \(Y_i \in \{0,1\}\) such as vote choice, one can apply a logistic link function:
\[
\hat{p}_i = \Pr(Y_i = 1 \mid x_i) 
= \sigma_{\text{logit}}\!\big(f(x_i)\big) 
= \frac{1}{1 + \exp\!\big(-f(x_i)\big)} .
\]
The loss function is the binary cross-entropy:
\[
\mathcal{L}(\theta) 
= - \frac{1}{N} \sum_{i=1}^N \Big[ y_i \log \hat{p}_i + (1-y_i)\log (1-\hat{p}_i) \Big],
\]
where \(\theta = \{W_\ell, b_\ell\}_{\ell=1}^L\).

For ordinal outcomes (e.g., Likert scales) or continuous variables (e.g., racial resentment index), 
I directly model \(Y_i \approx f(x_i)\) without a link function. 
In such cases, a standard regression loss such as mean squared error (MSE) is used:
\[
\mathcal{L}(\theta) = \frac{1}{N} \sum_{i=1}^N \big( y_i - f(x_i) \big)^2 .
\]
 
I decompose the network into a backbone and a task-specific head:
\[
f(x;\theta) = h_\psi \!\big( g_\phi(x) \big),
\]
where \(g_\phi\) is the backbone (shared feature extractor) and \(h_\psi\) is the task-specific head.  
I write \(\phi = (\phi^{\text{frozen}}, \phi^{\text{train}})\) where:
\begin{itemize}
  \item \(\phi^{\text{frozen}}\) are early-layer weights transferred from CES and kept fixed,  
  \item \(\phi^{\text{train}}\) are later-layer weights fine-tuned on ANES.
\end{itemize}

The transferred predictor is thus:
\[
\hat{y}_i = h_\psi \!\big( g_{\phi^{\text{train}}}( g_{\phi^{\text{frozen}}}(x_i)) \big).
\]

If \(\theta_S^\ast\) are weights learned on the CES source task, then the ANES fine-tuning stage initializes
\[
\theta_T^{(0)} = \Big( \phi^{\text{frozen}} \gets \phi_S^\ast,\;\; 
\phi^{\text{train}} \gets \phi_S^\ast,\;\; 
\psi \sim \text{random initialization} \Big),
\]
and only \((\phi^{\text{train}}, \psi)\) are updated during target-domain training.

\begin{figure}
    \centering
    \includegraphics[width=1\linewidth]{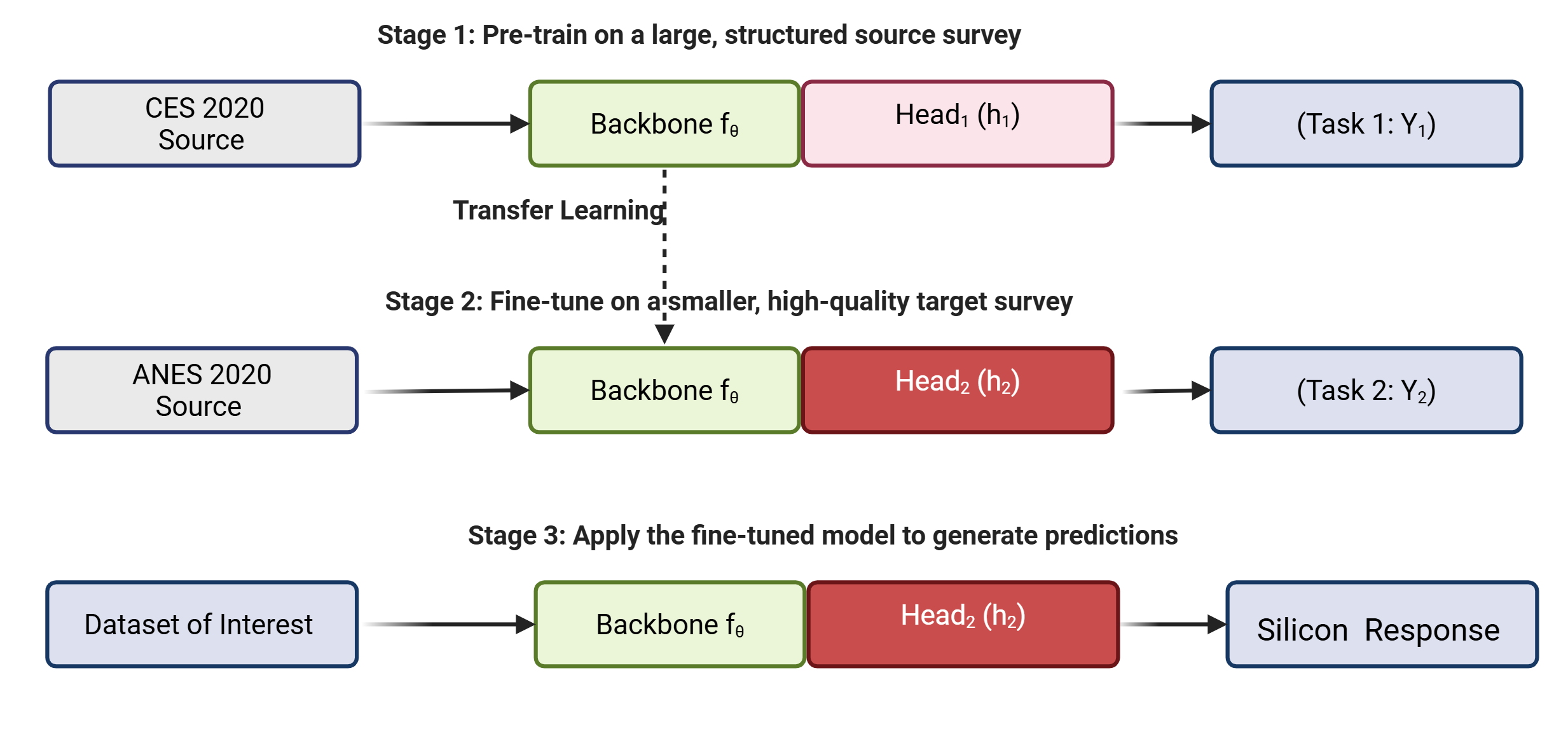}
    \caption{\textbf{Three-stage Survey Transfer Learning (STL) Framework.}
All stages share a common backbone \( f_{\theta} \), which encodes demographic and ideological representations (Anchor Transfer Variables, ATVs). 
\textbf{Stage 1:} The model is pre-trained on CES 2020 (Task~1: \(Y_1\)) using a backbone \( f_{\theta} \) and task-specific head \( h_1 \). 
\textbf{Stage 2:} The same backbone \( f_{\theta} \) is transferred by freezing its early layers to retain useful representations learned from CES data, while making the later layers and a new task-specific head \( h_2 \) trainable and fine-tuning them on ANES 2020 (Task~2: \(Y_2\)), yielding an updated backbone \( f_{\theta}^{\star} \).
\textbf{Stage 3:} The fine-tuned backbone \( f_{\theta}^{\star} \) and head \( h_2 \) are reused to generate empirically grounded silicon responses \(\hat{Y}_2\) for CES 2022 or any new dataset with shared features. 
This framework illustrates how knowledge can be transferred across surveys through a shared representation while adapting to new data distributions.}
    \label{fig:STLFlow}
\end{figure}

Figure 2 depicts the three-stage \textit{Survey Transfer Learning} (STL) pipeline, consisting of (1) pre-training on the source survey, (2) transfer and partial fine-tuning on the target survey, and (3) generating synthetic (``silicon”) responses on new data. The corresponding neural network architectures for each stage are summarized in Table~\ref{tab:stl_architecture}, which specifies the layer structure, activation functions, and trainable components. This design represents a basic backbone–head architecture, chosen deliberately for clarity and interpretability. While more complex architectures could potentially improve predictive performance, starting with this baseline highlights how even a simple transfer learning approach can yield substantial methodological benefits.

At the same time, I  emphasize that even this baseline architecture has useful theoretical properties relevant for survey applications in the social sciences. In particular, following recent theoretical scholarship from computer science, I employ ReLU activation functions in all hidden layers. ReLU networks are known to achieve minimax-optimal approximation rates in high-dimensional regression \citep{schmidt-hieber2020nonparametric}, with further elaboration in the rejoinder \citep{schmidt-hieber2020rejoinder}. Moreover,  \cite{ma2022theoretical} theoretical piece extends these guarantees to settings with temporally dependent data, which is especially pertinent for survey transfer learning where repeated cross-sections and panel designs may exhibit dependence across time.

\subsection*{Stage 1: Pre-training on the Source Domain}
A feedforward fully connected neural network \(M_S\) is trained on the source dataset \((X_S, Y_S)\). 
For binary outcomes (e.g., vote for Trump), I apply the logistic link function:
\[
\hat{p}_i = \Pr(Y_i = 1 \mid x_i) = \sigma_{\text{logit}}\!\big(f(x_i)\big),
\]
with binary cross-entropy loss:
\[
\theta_S^* = \arg\min_{\theta_S} \; 
\Bigg[- \frac{1}{N} \sum_{i=1}^N \big(y_i \log \hat{p}_i + (1-y_i)\log (1-\hat{p}_i)\big)\Bigg].
\]
For ordinal or continuous outcomes (e.g., racial resentment), the network directly models 
\(Y_i \approx f(x_i)\) with mean squared error loss. 
All backbone layers are trainable in this stage, and the model learns generalizable demographic–partisan representations.

\subsection*{Stage 2: Transfer and Fine-tuning on the Target Domain}
The learned weights \(\theta_S^*\) from CES pretraining initialize the target model on ANES \((X_T, Y_T)\). 
I decompose the parameters as \(\phi = (\phi^{\text{frozen}}, \phi^{\text{train}})\), where early layers 
\(\phi^{\text{frozen}}\) are fixed and later layers \(\phi^{\text{train}}\) are fine-tuned along with a new head \(h_2\):
\[
\theta_T^* = \arg\min_{\theta_T} L_T(Y_T, f(X_T; \theta_T)), \quad 
f(x;\theta_T) = h_2\!\big(g_{\phi^{\text{train}}}( g_{\phi^{\text{frozen}}}(x))\big).
\]
This preserves CES-informed demographic structure while adapting to ANES-specific distributions.

\subsection*{Stage 3: Generating Silicon Responses on New Data}
The tuned backbone \(f_\theta^\star\) is reused with a third head \(h_3\) to generate synthetic outcomes 
\(\hat{Y}_{\text{silicon}}\) on new data such as CES 2022:
\[
\hat{Y}_{\text{silicon}} = h_3\!\big(f_\theta^\star(x)\big).
\]
Because CES 2022 includes observed ground truth, I can validate predictions at both the individual level 
(accuracy, F1, AUC) and distributional level (e.g., Wasserstein distance, calibration). 
In this way, STL produces empirically grounded ``silicon responses'' and avoids the instability of LLM-based synthetic data.

\section*{Result and Discussion}

I report initial results using a basic deep learning architecture implemented; the available code can be run with any machine free in Google Colab. The implementation prioritizes computational accessibility and methodological transparency, requiring no specialized hardware or proprietary software. these findings demonstrate that deep learning architectures in general and survey transfer learning specifically  perform effectively for American public opinion survey data. This approach offers scholars a practical alternative to popular large language models, with high computational costs, very limited transparency, and concerns about reproducibility and carbon emission. As I demonstrate later in, STL also outperforms LLMs on sensitive political questions such as racial resentment measures, where nuanced demographic-partisan patterns are crucial for accurate prediction.

\subsection*{Predictive Performance (Individual-Level)}

\begin{figure}[H]
    \centering
    \begin{subfigure}[b]{0.48\textwidth}
        \centering
        \includegraphics[width=\linewidth]{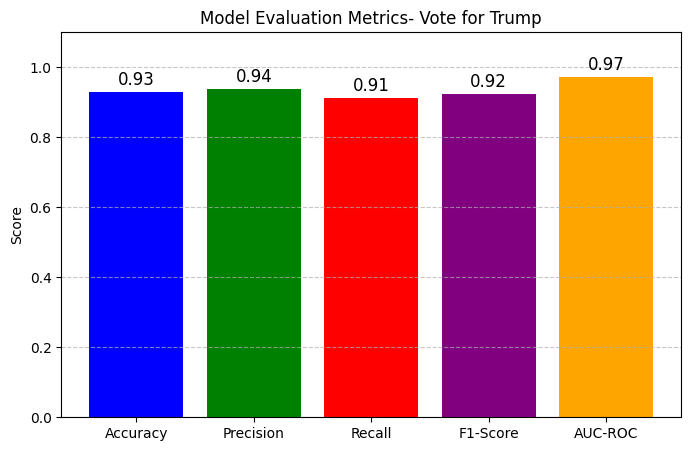}
        \caption{Vote for Trump (predicted vs. actual)}
        \label{fig:voter}
    \end{subfigure}
    \hfill
    \begin{subfigure}[b]{0.48\textwidth}
        \centering
        \includegraphics[width=\linewidth]{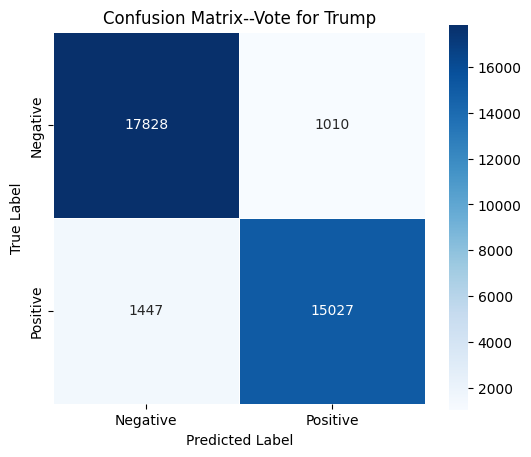}
        \caption{Confusion matrix for Vote for Trump}
        \label{fig:vote2}
    \end{subfigure}
    \caption{Model performance on vote choice prediction. (a) Aggregate Trump vote predictions compared with actual outcomes. (b) Confusion matrix showing class-level performance.}
    \label{fig:vote_results}
\end{figure}

As Figure~\ref{fig:vote2} and Figure~\ref{fig:voter} illustrate, the full STL cascade (CES 2020 → ANES 2020 → CES 2022) achieves strong performance, with an AUC of about 0.97. The AUC, or area under the receiver operating characteristic (ROC) curve, measures the model’s ability to distinguish between classes, with values closer to 1 indicating near-perfect discrimination. This demonstrates that by pretraining a simple deep learning model on CES and fine-tuning it on ANES, researchers can create accurate silicon responses for CES 2022, effectively producing a high-fidelity synthetic sample of vote choice. Focusing on the Vote for Trump outcome, a binary variable that is both central to American electoral politics and consistently measured with high accuracy in surveys, highlights the substantive utility of STL. It captures one of the most consequential political behaviors, with performance rivaling within-survey benchmarks.
\begin{table}[htbp]
\centering
\caption{Predictive performance across different survey-specific test sets.}
\label{tab:different_test}
\begin{tabular}{llccc}
\toprule
\textbf{Target} & \textbf{Model} & \textbf{Accuracy} & \textbf{F1 Score} & \textbf{AUC} \\
\midrule
\multirow{4}{*}{\textbf{Gun Ban}} 
 & Baseline ANES (80$\rightarrow$20)             & 0.700 & 0.809 & 0.782 \\
 & Baseline CES 22 (80$\rightarrow$20)           & 0.779 & 0.815 & 0.833 \\
 & Transfer CES 20 $\rightarrow$ ANES 20 (ANES 20\%) & 0.778 & 0.831 & 0.801 \\
 & Transfer CES 20 $\rightarrow$ ANES 20 $\rightarrow$ CES 22 & 0.779 & 0.823 & 0.840 \\
\midrule
\multirow{4}{*}{\textbf{Racial resentment 1}} 
 & Baseline ANES (80$\rightarrow$20)             & 0.779 & 0.678 & 0.819 \\
 & Baseline CES 22 (80$\rightarrow$20)           & 0.809 & 0.787 & 0.878 \\
 & Transfer CES 20 $\rightarrow$ ANES 20 (ANES 20\%) & 0.787 & 0.697 & 0.848 \\
 & Transfer CES 20 $\rightarrow$ ANES 20 $\rightarrow$ CES 22 & 0.825 & 0.792 & 0.877 \\
\midrule
\multirow{4}{*}{\textbf{Racial resentment 2}} 
 & Baseline ANES (80$\rightarrow$20)             & 0.762 & 0.697 & 0.806 \\
 & Baseline CES 22 (80$\rightarrow$20)           & 0.801 & 0.815 & 0.874 \\
 & Transfer CES 20 $\rightarrow$ ANES 20 (ANES 20\%) & 0.761 & 0.673 & 0.838 \\
 & Transfer CES 20 $\rightarrow$ ANES 20 $\rightarrow$ CES 22 & 0.806 & 0.802 & 0.871 \\
\bottomrule
\end{tabular}
\end{table}

\begin{table}[htbp]
\centering
\caption{Cross-survey transfer learning performance evaluated on a common CES 2022 test set.}
\label{tab:same_test}
\begin{tabular}{llccc}
\toprule
\textbf{Target} & \textbf{Model} & \textbf{Accuracy} & \textbf{F1 Score} & \textbf{AUC} \\
\midrule
\multirow{3}{*}{\textbf{Gun Ban}} 
 & CES 20 $\rightarrow$ CES 22 & 0.781 & 0.826 & 0.841 \\
 & ANES 20 $\rightarrow$ CES 22 & 0.740 & 0.814 & 0.818 \\
 & CES 20 $\rightarrow$ ANES 20 $\rightarrow$ CES 22 & 0.770 & 0.827 & 0.841 \\
\midrule
\multirow{3}{*}{\textbf{Racial resentment 1}} 
 & CES 20 $\rightarrow$ CES 22 & 0.826 & 0.798 & 0.883 \\
 & ANES 20 $\rightarrow$ CES 22 & 0.808 & 0.759 & 0.879 \\
 & CES 20 $\rightarrow$ ANES 20 $\rightarrow$ CES 22 & 0.823 & 0.790 & 0.882 \\
\midrule
\multirow{3}{*}{\textbf{Racial resentment 2}} 
 & CES 20 $\rightarrow$ CES 22 & 0.807 & 0.805 & 0.871 \\
 & ANES 20 $\rightarrow$ CES 22 & 0.803 & 0.806 & 0.866 \\
 & CES 20 $\rightarrow$ ANES 20 $\rightarrow$ CES 22 & 0.804 & 0.798 & 0.870 \\
\bottomrule
\end{tabular}
\end{table}

As shown in Table 1 and Table 2, even this proof-of-concept implementation demonstrates that transfer learning can reproduce survey responses with striking accuracy across datasets. In Table 1, models trained and tested on different survey waves (CES 2020, ANES 2020, and CES 2022) achieve high accuracy and AUC, confirming that deep neural networks can recover meaningful attitudinal patterns within and across surveys. Table 2 further shows that when all models are evaluated on a common test set (CES 2022), cross-survey transfers remain robust, indicating that learned representations generalize beyond their original sampling frames. These results highlight the potential of Survey Transfer Learning (STL) as both a validation and imputation framework for major surveys such as ANES and CES. Future studies can refine these models through larger architectures, additional regularization strategies, and broader feature spaces, but the current results establish a clear methodological proof of concept for the approach.

For scholars, this approach\footnote{This approach differs fundamentally from recent attempts at AI-generated synthetic data. While studies like \cite{argyle2023out} have explored LLMs' ability to generate "silicon samples" conditioned on demographic profiles, these models struggle with higher-order relationships and statistical precision. \cite{bisbee2024synthetic} found that 48\% of regression coefficients from LLM-generated responses diverged substantially from ANES data, with 32\% showing directional changes in effect size. Moreover, LLM-generated responses often fail basic replication tests due to algorithmic instability \citep{Rossi2024}.
} shows that ANES is not only a high-quality source of data but also a training ground for methodological innovation. Using STL, researchers can test hypotheses about political behavior with silicon responses that are empirically grounded, reproducible, and validated against held-out cases. In this sense, ANES can serve simultaneously as both data provider and methodological benchmark for the next generation of survey research.

\begin{figure}
    \centering
    \includegraphics[width=1\linewidth]{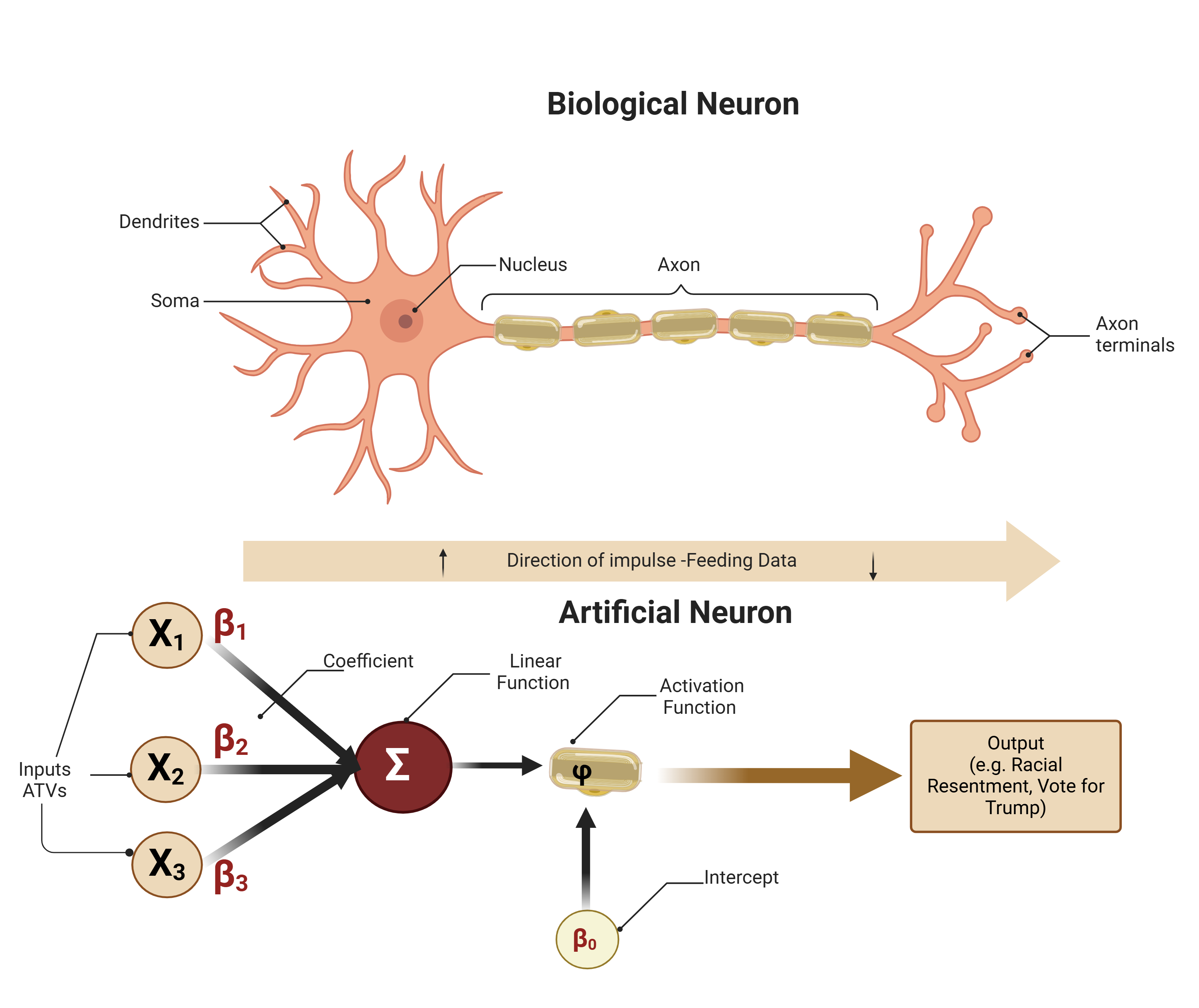}
    \caption{Comparison of a biological neuron and an artificial neuron used in deep learning.}
    \label{fig:neuron}
\end{figure}

Baseline results provide a proof of concept that transfer learning improves upon single-survey models, showing that knowledge can be transferred between surveys to enhance predictive accuracy. More broadly, the consistently high accuracy achieved with a simple deep learning model and limited features demonstrates that such models are well-suited for capturing public opinion patterns in large-scale survey data.These binary outcome results validate this framework regarding the relationship between political polarization, sorting, and demographic predictability \citep{bishop_2008, hetherington_2009}.

Figure~\ref{fig:neuron} helps explain why this works. Artificial neurons process inputs (demographics, partisanship, and other anchor transfer variables) through a linear combination of coefficients and then apply a nonlinear activation function to produce outputs such as racial resentment or vote choice. In this sense, artificial neurons are structurally similar to linear models, but with the added capacity to approximate complex, high-dimensional functions. This capacity allows STL to capture richer patterns than traditional regression and explains why it can achieve consistently high predictive validity across surveys. As shown in Supplemental Material 2, STL can be broadly seen as a solution to the missing data problem as well, outperforming multiple imputation which achieves 78\% accuracy. While multiple imputation relies on assumptions about the missing data scenarios, STL operates as a prediction framework without requiring these parametric assumptions.

Thus, these findings show that STL offers a more empirically grounded and reproducible approach to synthetic data generation than large language models. While LLMs often hallucinate or produce biased outputs, STL relies on real survey distributions, transfers knowledge transparently, and validates performance against held-out respondents. This combination of interpretability and empirical rigor makes STL a stronger foundation for survey methodology going forward.

\begin{figure}
    \centering
    \includegraphics[width=1\linewidth]{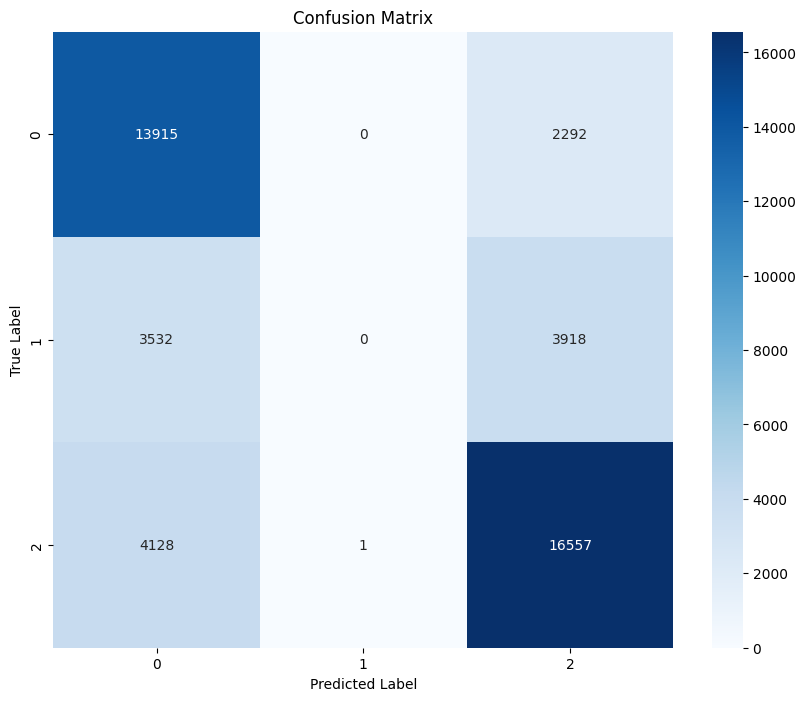}
    \caption{The confusion martix.}
    \label{racial resentment3}
\end{figure}

The confusion matrix in Figure \ref{racial resentment3} reveals that the model achieves high accuracy for respondents with clear ideological positions about racial resentment questions but struggles significantly with neutral responses on attitude items. This pattern suggests that ``neither agree nor disagree'' responses represent fundamentally different response processes, potentially due to social desirability bias, satisficing behavior, or genuine ambivalence, that are less predictable from standard demographic and political covariates. This finding motivates the binary recoding of attitude scales for STL at this stage. Future studies can extend this framework to work with ordinal and continuous outcomes as modeling techniques and sample sizes improve.

For binary outcomes, STL demonstrates measurable improvements over single-survey baselines. Note that I collapsed the original five-point ordinal scales of racial resentment to binary outcomes and used casewise deletion for missing data. On ANES holdout data, pretraining on CES 2020 before fine-tuning on ANES 2020 improves AUC performance for both outcomes compared to ANES-only models. When applying the full cascade to predict CES 2022 outcomes, the transfer learning approach maintains competitive performance, matching or slightly exceeding CES-only baselines despite the additional complexity of cross-survey transfer.

These results provide initial evidence that demographic-partisan representations learned from one survey context can transfer to others for attitudinal measures beyond simple vote choice. While the performance gains are modest, they suggest that STL can leverage existing survey data to generate predictions in new contexts without substantial loss of accuracy, a valuable property for cost-effective survey research and data integration efforts.\footnote{These individual-level predictions have important implications for distributional survey research. Many survey users require accurate distributional estimates rather than individual-level predictions, and some organizations now resample data to preserve distributional accuracy while protecting privacy. High individual-level accuracy enables generation of synthetic responses that preserve population-level patterns while maintaining anonymity. Future studies can extend STL to polling applications and continuous variables using richer feature sets, supporting split-sample validation, nonresponse bias diagnostics, and cross-survey harmonization while preserving distributional properties.}

\subsection*{Distribution-Level Validation}
\begin{figure}[htbp]
    \centering
    \begin{subfigure}[b]{0.48\linewidth}
        \centering
        \includegraphics[width=\linewidth]{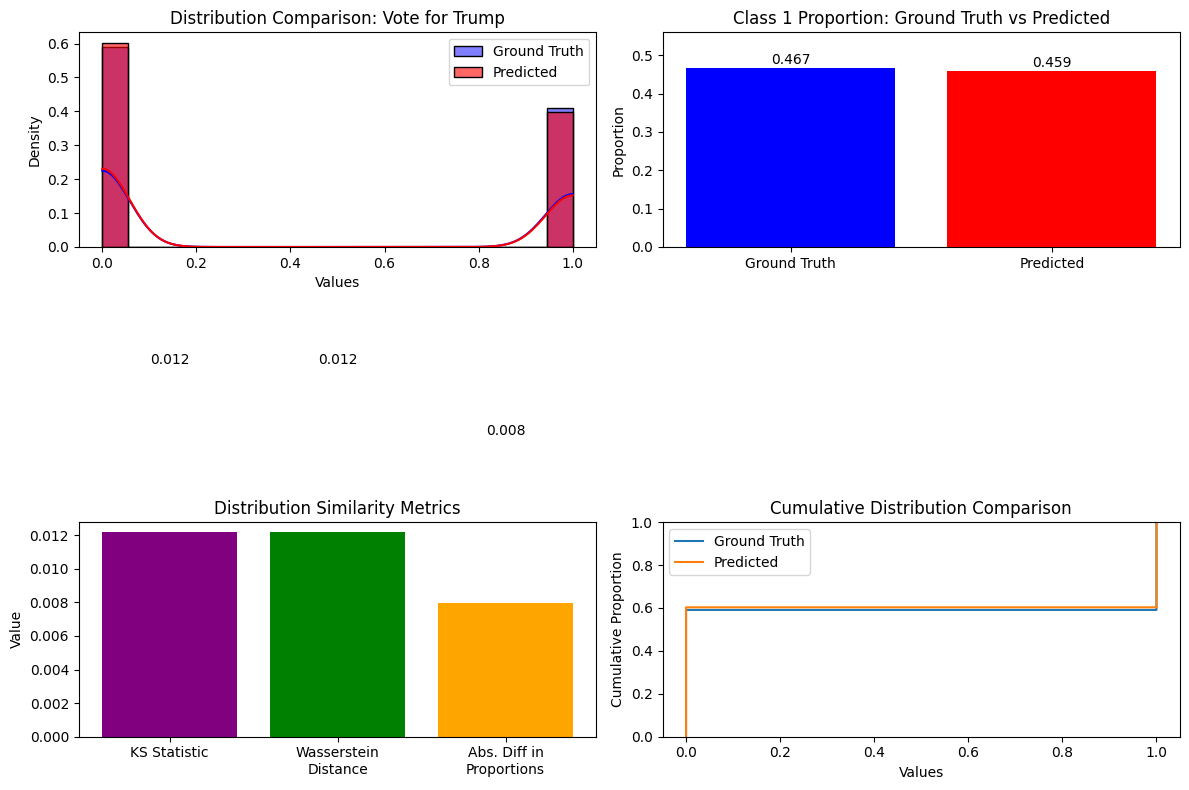}
        \caption{Trump support (CES 2022)}
        \label{fig:dis}
    \end{subfigure}
    \hfill
    \begin{subfigure}[b]{0.48\linewidth}
        \centering
        \includegraphics[width=\linewidth]{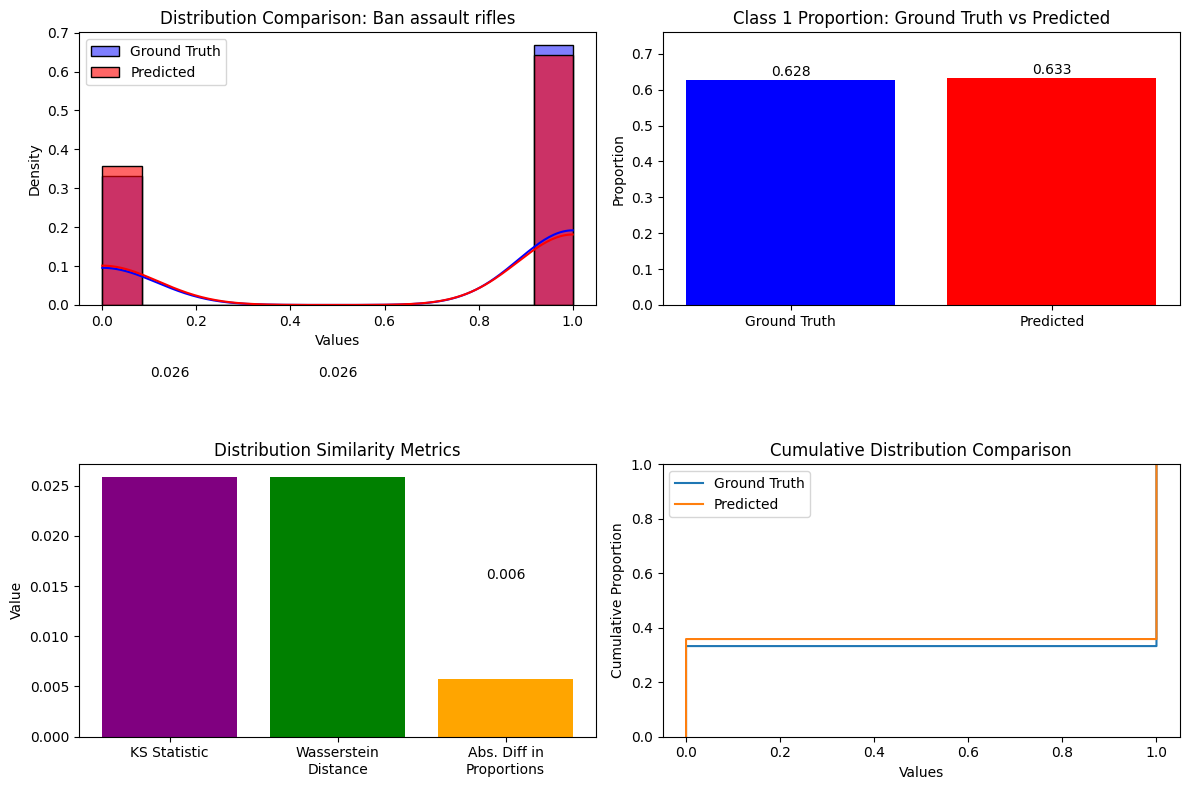}
        \caption{Ban assault rifles (ANES 2020)}
        \label{fig:gundisb}
    \end{subfigure}
    
    \begin{subfigure}[b]{0.48\linewidth}
        \centering
        \includegraphics[width=\linewidth]{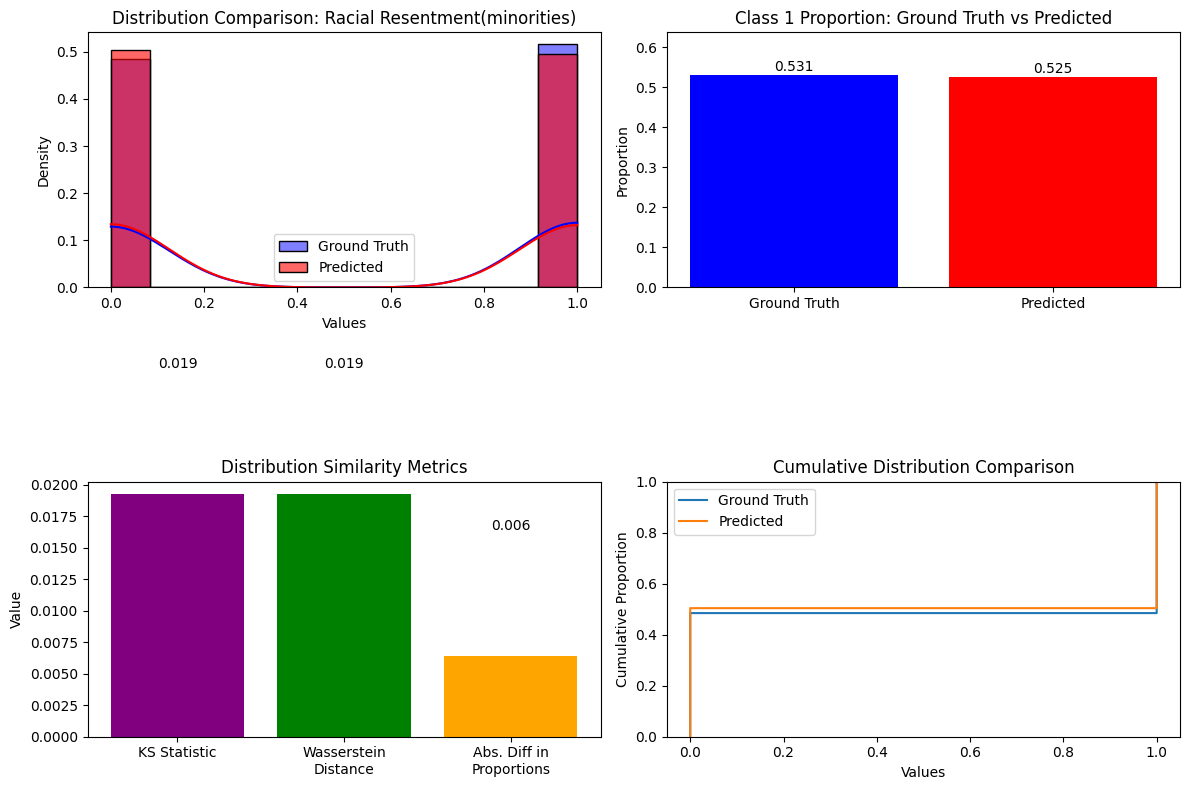}
        \caption{Racial resentment (ANES 2020)}
        \label{fig:racial_resentment}
    \end{subfigure}
    \hfill
    \begin{subfigure}[b]{0.48\linewidth}
        \centering
        \includegraphics[width=\linewidth]{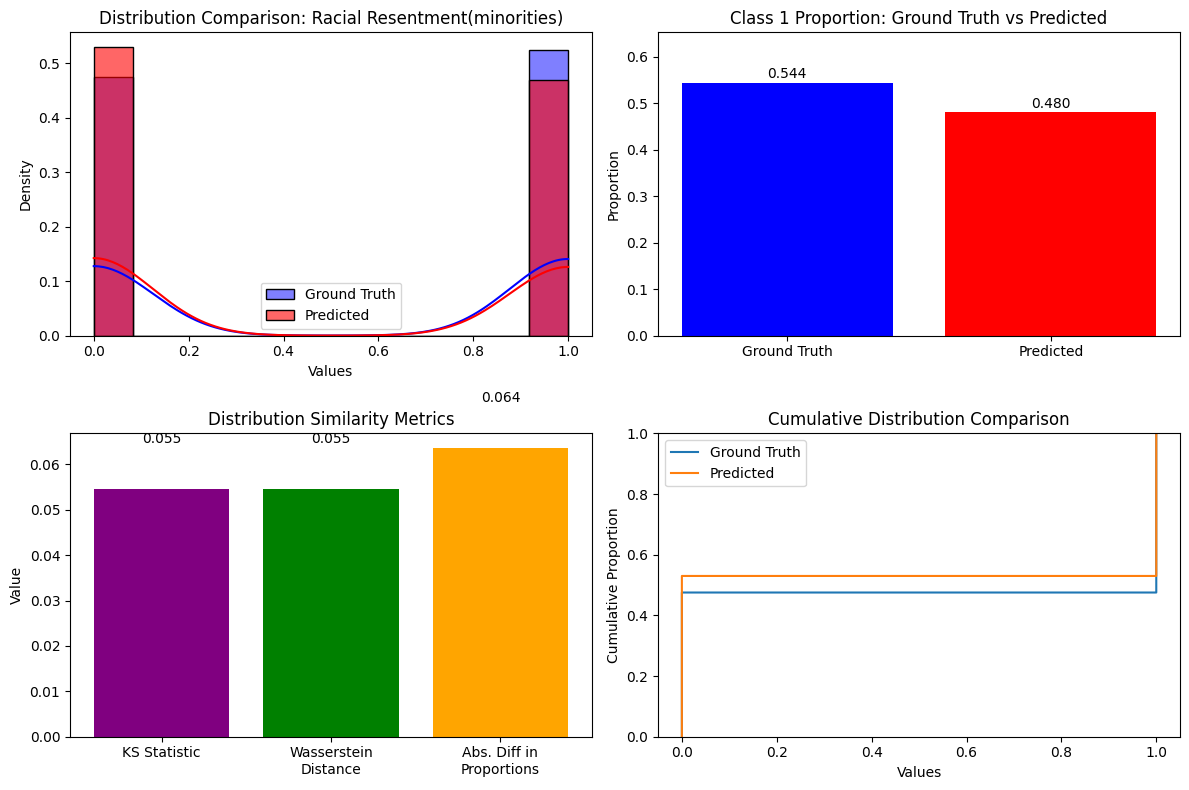}
        \caption{Racial resentment, minorities (ANES 2020)}
        \label{fig:racial_resentment_minorities}
    \end{subfigure}
    
    \caption{Comparison of actual and model-predicted distributions across selected survey items.}
    \label{fig:distribution_comparisons}
\end{figure}
Beyond individual silcin response, STL reproduces aggregate distributions with high fidelity.  
Table~\ref{tab:dist_metrics_summary} summarizes results for Trump vote, assault rifle bans, and racial resentment.  
In each case, predicted distributions closely match observed ones: KS statistics are $<0.03$, Wasserstein distances $<0.03$, and absolute proportion errors $<1$ percentage points.  
For the minority resentment subgroup, divergence is somewhat larger (KS = 0.055), highlighting a case where further methodological refinement is needed. High-fidelity replication of aggregate distributions suggests STL can support applications in forecasting, survey calibration, and missing-data recovery.

\begin{table}[htbp]
\centering
\caption{Distribution-Level Comparison Across Outcomes}
\begin{tabular}{lccc}
\hline
\textbf{Outcome} & \textbf{KS Stat. (p)} & \textbf{Wass. Dist.} & \textbf{Abs. Diff.} \\
\hline
Trump vote (CES22) & 0.012 (p=0.87) & 0.012 & 0.009 \\
Ban assault rifles (ANES20) & 0.026 (p=0.85) & 0.026 & 0.006 \\
Racial resentment (ANES20) & 0.025 (p=0.85) & 0.025 & 0.025 \\
Minority resentment (ANES20) & 0.055 (p=0.08) & 0.055 & 0.064 \\
\hline
\end{tabular}
\label{tab:dist_metrics_summary}
\end{table}

\subsection*{Silicon Responses as an Empirical Alternative to LLM Hallucinations}

Is LLM Better than STL in Practice? As discussed throughout this paper, there are substantial concerns with using LLMs to generate synthetic survey responses or an entire ``silicon sample''. Most notably, these models are trained on unstructured internet text rather than empirical national representative survey data like ANES or CES, resulting in synthetic outputs that often lack empirical validity and deviate from the statistical properties of real survey responses \citep{Rossi2024, bisbee2024synthetic}. I argue that scholars often do not need a full 'silicon sample' since they already have ATV questions in their survey, but 'silicon response' which can be transfer from other survey.

LLMs such as GPT-3 and GPT-4 frequently suffer from \textit{hallucinations}, outputs that are syntactically plausible but factually incorrect, as well as ideological and demographic biases inherited from their pretraining corpora \citep{mckenna-etal-2023-sources, weidinger2022taxonomy}. These models are fundamentally opaque: their training data is inaccessible, their architectures proprietary, and their outputs non-replicable \citep{agnew2024illusion}. Even scholars who have published studies using GPT-3 rarely have access to the model itself, relying instead on APIs that limit transparency and control. As a result, synthetic survey responses generated by LLMs can vary significantly across runs, undermining scientific reproducibility and raising serious ethical and epistemological concerns \citep{Rossi2024, Pilati2024, pokotylo2024ethical}.

Moreover, the cost of deploying LLMs at scale is prohibitively high, both computationally and financially. These models require significant GPU resources which could be concerning for environments \citep{Ren2024LLMEnvironment}, and the human cost of validation—especially for labeling and checking outputs across diverse demographic subgroups, is substantial and often overlooked in methodological discussions.

However, the underlying architecture of LLMs, the deep learning foundations that make them powerful, can be repurposed more effectively and directly for survey research through Survey Transfer Learning (STL). STL retains the strengths of deep learning while avoiding the drawbacks of LLM-generated synthetic data. Instead of generating text-based responses from scratch, STL relies on actual observed responses from high-quality surveys and transfers learned patterns between datasets using shared features like demographics and party ID (Anchor Transfer Variables, or ATVs).

\subsubsection*{LLM Predictions on the Racial Resentment}

\begin{figure}
    \centering
    \includegraphics[width=1\linewidth]{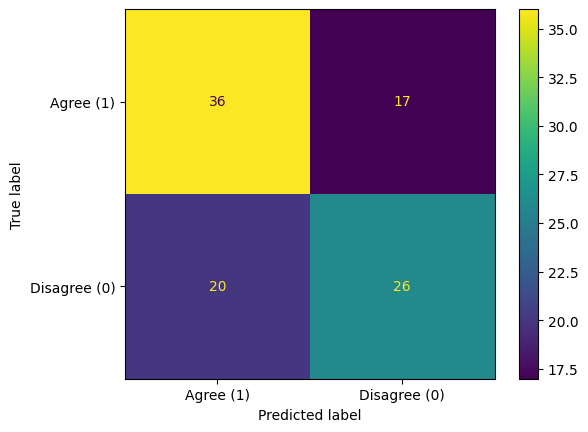}
    \caption{Confusion matrix for LLM predictions on the racial resentment item. The model shows higher recall for the \textit{Agree} class but struggles with accurate classification of \textit{Disagree} responses.}
    \label{fig:Confusion}
\end{figure}
As the confusion matrix \ref{fig:Confusion} shows, the model achieved 62\% accuracy overall, with a precision of 64\% and a recall of 68\% for the \textit{``Agree''} class. However, performance dropped notably for the \textit{``Disagree''} class (precision = 60\%, recall = 56\%), indicating particular difficulty in recognizing when respondents would reject the statement.

These results are consistent with the emerging evidence that LLMs suffer from structural biases and hallucinations when used for synthetic response generation on sensitive topics. The model's inability to capture the moral and political nuance embedded in this subtle racial resentment measure shows key  limitations in understanding how demographic characteristics shape attitudes toward race and inequality.

\begin{figure}
    \centering
    \includegraphics[width=1\linewidth]{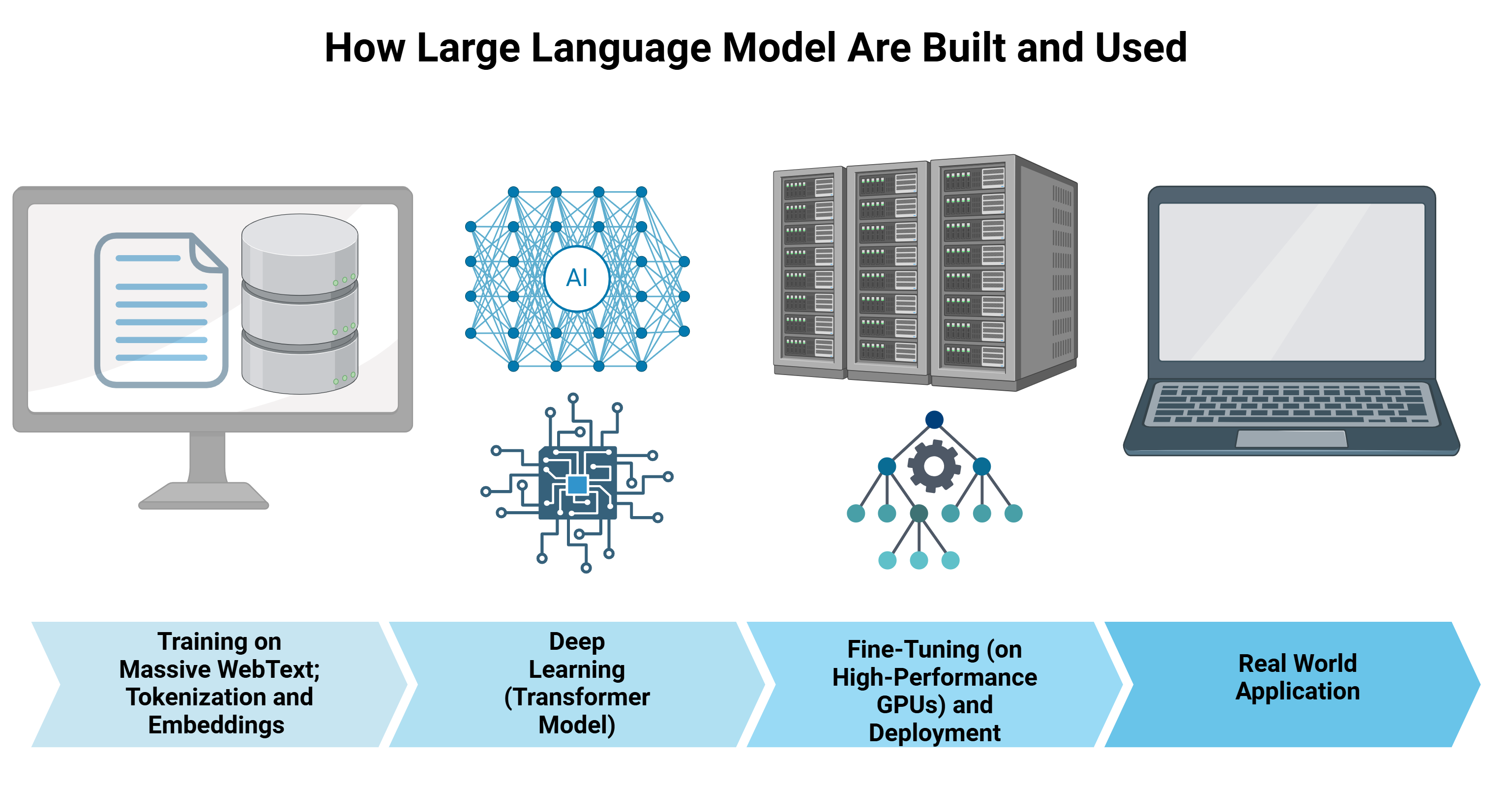}
    \caption{The four key stages in the development and use of large language models. Importantly, LLMs are  trained on massive web-based text corpora. These texts are first converted into numerical representations through a process known as tokenization and embedding. Next, deep learning—typically using transformer architectures, enables the model to learn statistical patterns in the data. In the third stage, the model is fine-tuned for specific tasks using high-performance GPUs and then deployed. Finally, the model is integrated into real-world applications where it generates responses based on user input.}
    \label{fig:LLM}
\end{figure}

Finally, as  Figure~\ref{fig:LLM} illustrates the contrast: LLMs are trained on unstructured internet text, while STL builds on gold-standard survey data.  
In practice, LLMs systematically misclassify subtle racial resentment items, over-predicting ``Agree” responses.  
As shown above, accuracy is about 60\%, with especially poor recall for the \textit{Disagree} class.  

By contrast, STL achieves over 90\% accuracy on the same outcomes and replicates distributions within one percentage point.  
STL also provides transparency (open data, reproducible models) and efficiency (trainable in commodity notebooks), avoiding the opacity, instability, and high computational cost of LLMs.  
\section*{Conclusion}

This paper introduces Survey Transfer Learning (STL) and demonstrates that it is more accurate, accessible, and cost-effective than LLM-generated silicon responses. STL is especially well-suited to this historical moment, where polarization has increased the predictive power of demographic variables, we have unprecedented access to computational resources, and we face mounting challenges in traditional survey research. While political methodology societies have focused on LLMs as part of artificial intelligence, I call for utilizing other aspects of AI. Specifically, I advocate for the direct application of deep learning algorithms and transfer learning paradigms to surveys as structured data, given the methodological, environmental, and ethical constraints of LLMs, which are better suited for text-based applications.

I demonstrate that by pretraining a basic deep learning model on CES 2020 and fine-tuning on ANES 2020, we can transfer learning to create ``silicon respondents'' for CES 2022 or 20\% held-out data from ANES 2020. This serves as proof of concept, and future research can apply this framework to address critical issues in survey research. Future studies can explore this method with panel data, continuous variables, and imputed data. STL shows particular promise for surveys with limited observations, such as studies of Native American populations, immigrant communities, or other hard-to-reach groups. Additionally, STL could address survey nonresponse bias by leveraging patterns from high-response-rate surveys, harmonize questions across different survey instruments to enable cross-survey comparisons, and reduce costs associated with oversampling rare populations. The framework may also prove valuable for forecasting opinion dynamics by transferring knowledge from historical survey data to predict trends with minimal new data collection, and for addressing mode effects when surveys transition between telephone, online, and mixed-mode designs.

STL can help students of political science engage in data integration between currently available high-quality datasets from major surveys or repositories such as Harvard Dataverse, effectively recycling data instead of paying firms to run additional surveys or add additional questions with low quality or power. It can also be used for proxy finding and for resampling surveys for privacy concerns, as it shows remarkable distribution levels. 

Drawing on theories of political behavior and using advances in machine learning and Colab free computational space, I show how STL makes the most of limited data with the use of Anchor Transfer Variables (ATVs) to bridge between surveys that are traditionally analyzed separately. Treating surveys as interconnected rather than isolated sources of data via deep learning can help overcome limitations that have traditionally stemmed from computational constraints. In the context of American public opinion, I use shared demographic and ideological variables as ATVs, enabling reliable knowledge transfer between surveys like ANES and CES.  

The findings show that STL, even when implemented with a simple feedforward neural network, performs better than both multiple imputation and LLM-generated synthetic data across a range of tasks. Unlike LLMs, which often produce biased or unstable outputs, STL consistently avoids issues such as reproducing racist or sexist content. In particular, LLMs struggle with items on racial resentment, while STL maintains high predictive accuracy for binary outcomes such as vote choice, support for a gun ban, and related measures. For more complex ordinal outcomes like racial resentment scales, STL produces acceptable results, though accuracy is lower in neutral response categories. This is an expected limitation given the nature of missing-not-at-random responses. The results also provide proof of concept that STL can successfully transfer information between surveys in ways that are difficult to achieve with standard deep learning.

Beyond these immediate results, STL encourages a shift in how surveys are understood in the context of AI. Rather than viewing each survey as an isolated instrument, STL treats them as interconnected data sources that can be integrated using deep learning methods. In this perspective, demographic and ideological variables are not only descriptive but can also be transferable, opening opportunities for hypothesis testing, poll integration, and other methodological innovations.

\section{Appendix A: Survey Transfer Learning and Missing Data}

\begin{figure} [H]
    \centering
    \includegraphics[width=1\linewidth]{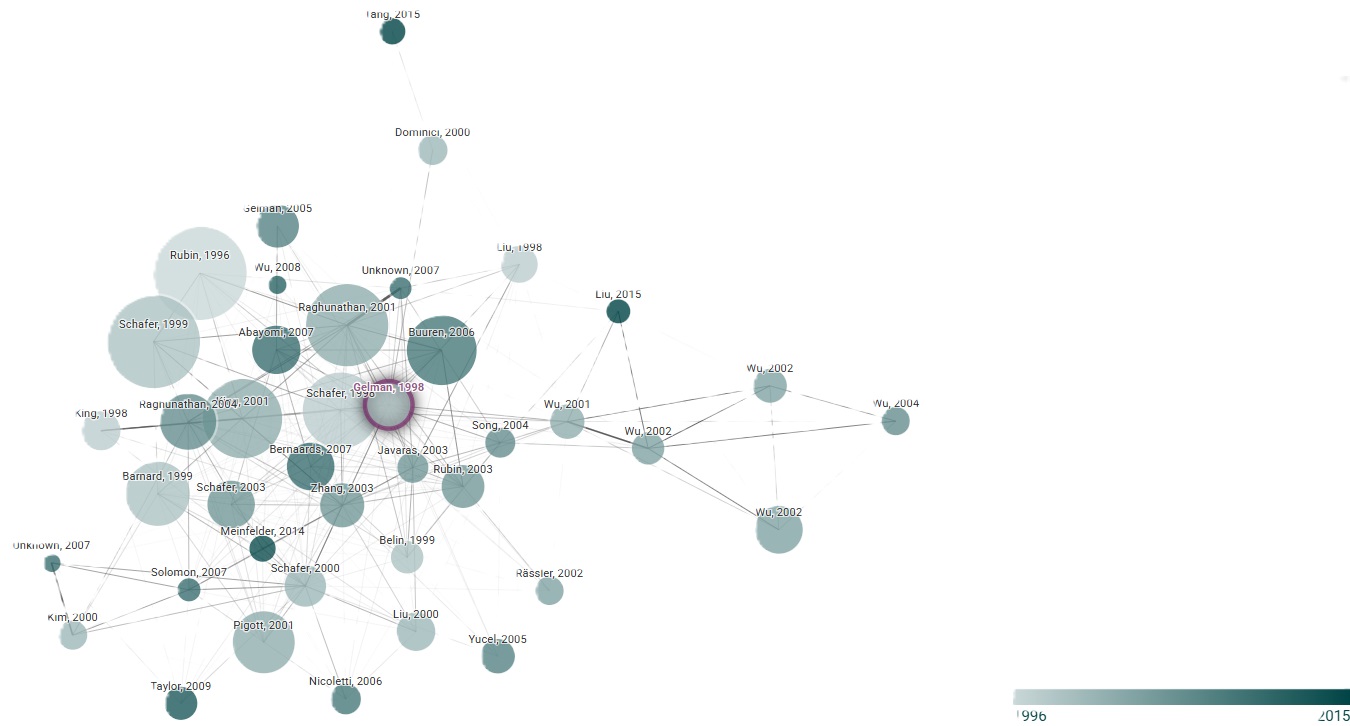}
    \caption{The paper by Andrew Gelman, Gary King, and Chuanhai Liu has received modest citation over the last three decades, with only a handful of empirical application. This suggests that either the paper did not find its audience, or vice versa. (Image from \cite{connectedpapers_not_asked_2025}).}
    \label{fig:enter-label}
\end{figure}

Nearly three decades ago, in \textit{ Not Asked and Not Answered: Multiple Imputation for Multiple Surveys} \citep{gelman1999notasked} introduced an approach for analyzing a series of independent cross-sectional surveys to address missing items and questions across multiple datasets. Their method extends existing imputation techniques designed for single surveys by incorporating a hierarchical regression model with Bayesian inference to estimate missing responses, effectively pooling information across datasets. While conceptually significant, their work had only a modest citation impact (around 150) and saw limited direct empirical applications, especially when compared to the nearly 5,300 citations of the papers that cited their work \citep{Rubin1996, Schafer1999}.

While their approach was to ``comprise fitting separate imputation models for each survey'' to address problems of imputing missing data from several datasets, we argue that a broad version of this underappreciated idea can be leveraged for generating synthetic responses, particularly through artificial intelligence frameworks and theories of political behavior. Therefore, while the authors were initially unaware of Gelman's article, this paper broadly revisits \cite{gelman1999notasked}'s idea, leveraging modern AI tools and insights from political behavior theory to generate synthetic responses with higher accuracy at the individual level compared to both multiple imputation (78\%) and LLM-generated data.

Specifically, I argue that STL can be considered as an emerging alternative for both multiple imputation \citep{Rubin1987Multiple} and LLM generating responses to unasked survey questions using deep learning architectures. Unlike multiple imputation, which assumes missing-at-random patterns and loses approximately 15\% accuracy with complex or large datasets, STL applies machine learning models pre-trained on one survey dataset (e.g., CES) and fine-tuned on another (e.g., ANES) to predict missing responses with more stable, reproducible outputs and near-empirical accuracy.

It is worth noting that even what I proposed to compare STL to—multiple imputation which binds the rows of two nationally representative surveys with the same additional treatment variables (ATVs) and then runs imputation approaches such as mice \citep{mice}—is an innovative usage of multiple imputation for generating synthetic data.

As summarized in Table~\ref{tab:stl_vs_llm}, STL also outperforms LLM-based synthetic data generation in the context of political surveys. While LLMs lack access to high-quality, structured survey datasets such as ANES or CES and instead rely on unstructured internet text corpora \citep{radford2018gpt1}, STL leverages these gold-standard datasets directly through supervised training and fine-tuning. This leads to more accurate, stable, and empirically grounded outputs, avoiding the hallucinations and ideological biases often observed in LLM-generated responses.

Table \ref{tab:stl_vs_llm} summarizes the fundamental differences between STL, LLM-generated synthetic survey data, and Multiple Imputation, highlighting STL's advantages while acknowledging its limitations.

\begin{table}[h]
\centering
\caption{Comprehensive Comparison of Methods for Survey Data Augmentation}
\label{tab:stl_vs_llm}
\begin{tabularx}{\textwidth}{|X|X|X|X|}
\hline
\textbf{Issue} & \textbf{Survey Transfer Learning (STL)} & \textbf{LLM-Generated Synthetic Data} & \textbf{Multiple Imputation} \\
\hline
\textbf{Data Foundation} & Uses real survey data (CES/ANES), maintaining empirical validity. & Inherits ideological biases from internet text corpora, lacking empirical survey data political responses. & Relies on existing data patterns, struggling with complex relationships. \\
\hline
\textbf{Political Distribution} & Preserves real distribution of survey responses. & Reflects polarization of pretraining corpora, skewing synthetic responses. & Assumes missing-at-random (MAR), potentially missing systematic patterns. \\
\hline
\textbf{Stability \& Replicability} & Produces stable, reproducible outputs with consistent performance metrics. & Responses vary randomly across runs, undermining scientific rigor. & Relatively stable but sensitive to model specification. \\
\hline
\textbf{Accuracy} & Near-empirical accuracy with predictable error margins. & Variable accuracy, highly context-dependent. & Approximately 15\% reduced accuracy compared to complete data analysis. \\
\hline
\textbf{Scalability} & Efficiently handles large survey datasets and complex variable relationships. & Handles complex data but computationally expensive. & Performance deteriorates with high-dimensional data or large missing portions. \\
\hline
\textbf{Ethical Considerations} & Grounded in real-world survey methodology with transparent assumptions. & Risks misrepresenting public opinion due to uncalibrated biases in training data. & May reinforce existing patterns of missingness, especially for marginalized groups. \\
\hline
\textbf{Computational Requirements} & Moderate computational needs. & Substantial computational resources required. & Relatively low computational demands but iterative nature can be time-consuming. \\
\hline
\end{tabularx}
\end{table}

\section{Appendix B: From Neurons to Surveys: A Deep Learning Framework for Social Science}

\begin{figure}
    \centering
    \includegraphics[width=1\linewidth]{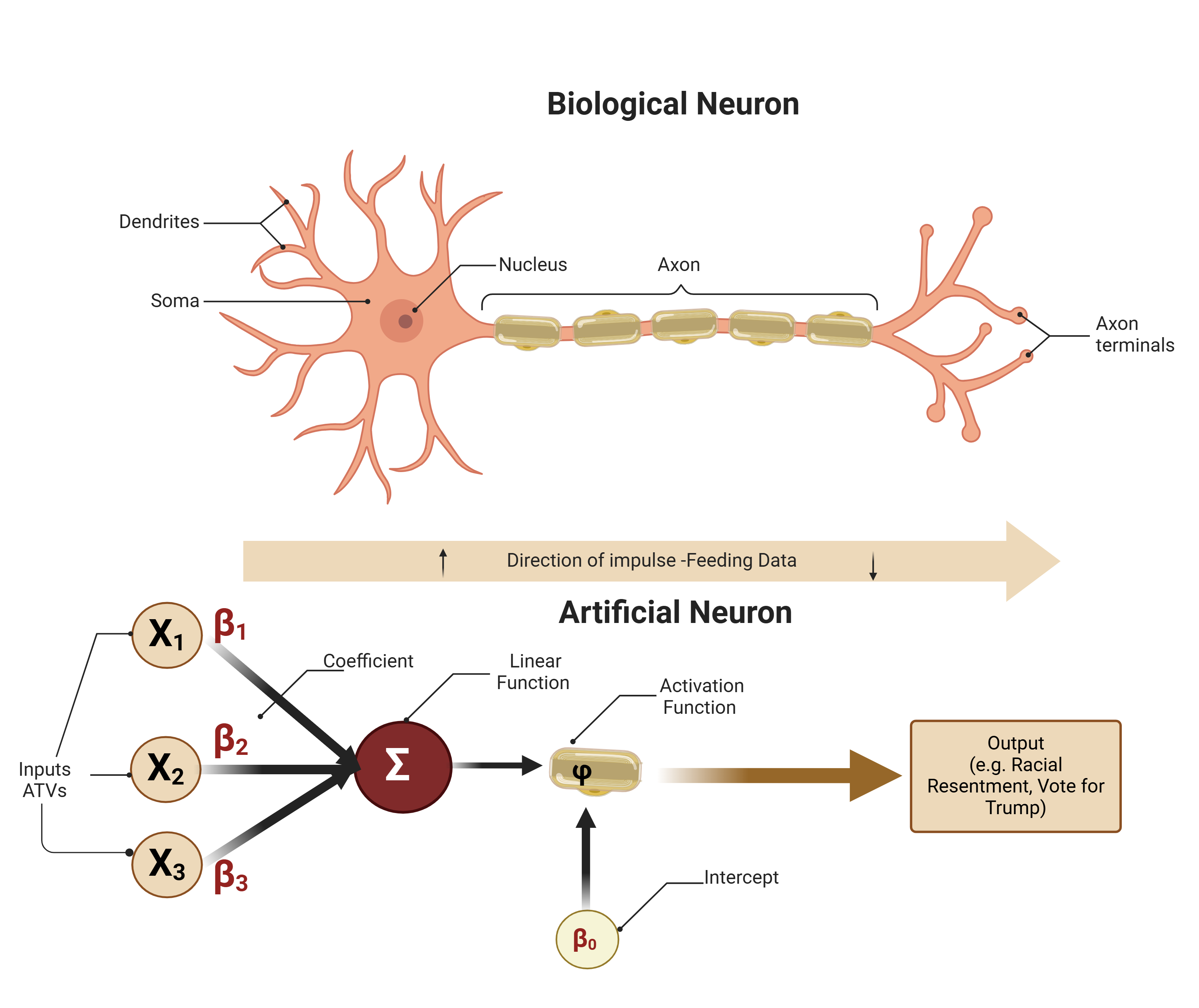}
    \caption{Comparison of a biological neuron and an artificial neuron used in deep learning.}
    \label{fig:Neuron}
\end{figure}

I use deep learning, a class of machine learning models that learn hierarchical representations by utilizing multiple layers of computation. At its core, deep learning is built on the concept of the artificial neuron, or perceptron, which is loosely inspired by the structure and function of biological neurons (see Figure~\ref{fig:Neuron}). In this architecture, input values \( x_i \) (in our case, shared ATV variables) are multiplied by corresponding weights \( w_i \) (what social scientists call coefficients), summed, and passed through a bias term \( b \) (intercept). The result is then transformed by a nonlinear activation function \( \phi \), allowing the model to capture complex, nonlinear relationships in the data. This structure of deep learning is inherently superior to multiple imputation, as the latter is unable to capture the complex, nonlinear relationships present in the data.

\begin{figure}
    \centering
    \includegraphics[width=1\linewidth]{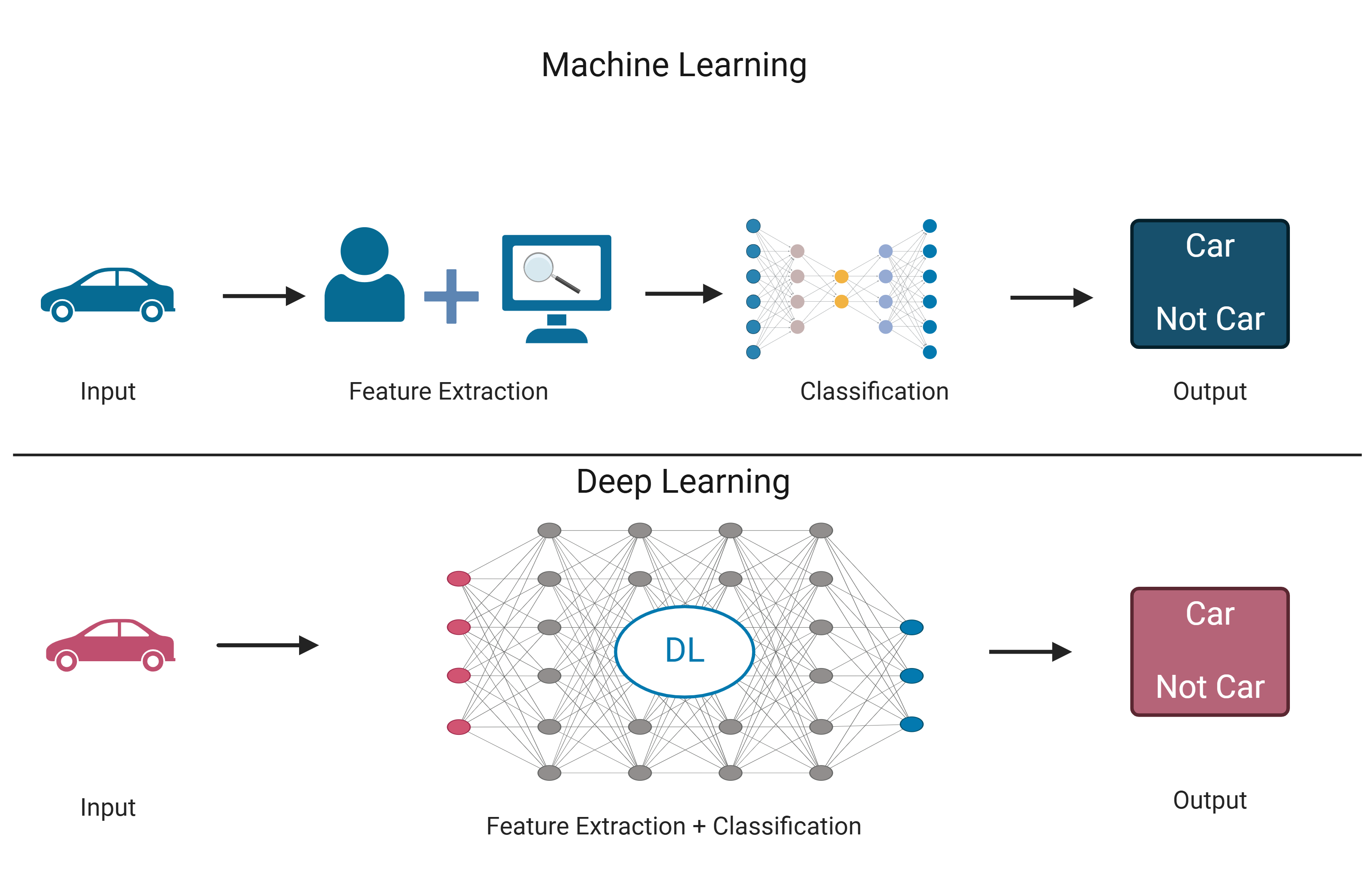}
    \caption{Traditional Machine Learning requires human-engineered feature extraction, where domain experts manually identify and extract relevant features from raw data before classification occurs. In contrast, Deep Learning integrates feature extraction and classification into a unified end-to-end learning process, automatically discovering hierarchical representations from raw input without explicit human intervention, enabling more sophisticated pattern recognition particularly for complex unstructured data like images, audio, and text.}
    \label{fig:dl_vs_ml}
\end{figure}

Unlike traditional machine learning approaches that often rely on hand-engineered features and separate feature extraction steps, deep learning automatically discovers and optimizes feature representations through end-to-end learning (see Figure~\ref{fig:dl_vs_ml}) \citep{Goodfellow2016}. In STL application, deep learning allows the model to learn from survey data with minimal manual preprocessing, making it both scalable and adaptable to new survey environments.

I specifically employ a feedforward neural network architecture as the backbone of the transfer learning framework, implemented using TensorFlow and Keras \citep{abadi2016tensorflow}.

Transfer learning is a machine learning ``paradigm'' that leverages previously trained models for a different but related task \citep{chao2023editorial, pan2010survey}. A canonical example of transfer learning is in driving: experienced passenger vehicle operators typically demonstrate accelerated acquisition of commercial transport skills. This transfer of domain knowledge to a different but related task shows how expertise in a source domain (passenger vehicle operation) can expedite mastery of a related target domain (commercial transport), despite differences in vehicle scale and operational complexity.

Researchers who prefer not to run new surveys—or who seek to enhance the value of existing datasets—can use Survey Transfer Learning (STL) as a practical and sustainable alternative. For example, a researcher working with a high-quality dataset that lacks a specific variable (e.g., policy preference) can identify another public survey that includes both the missing variable and overlapping ATVs. Using transfer learning principles, a model is first pre-trained on the source dataset (often larger N and lower quality data) and then transferred to the target (smaller N and higher quality) dataset. During this process, earlier model layers (which capture general patterns) can be frozen to preserve learned relationships, while later layers are fine-tuned using any available data in the target survey—enhancing performance without overfitting. This enables accurate prediction of missing responses even with limited data. STL allows researchers to test nuanced hypotheses and model complex interactions without launching a new survey. It can also be applied to original surveys: by asking only core ATV questions, researchers can use STL to recover unasked outcomes, reducing respondent burden while maintaining analytical rigor and interpretability.

\section{Appendix C: Outcomes}
The features, or as we call them in the paper, predictors, are gender, race, region, age, party ID, and education. I removed ideology as it has a 0.75 correlation with party ID and has 3,500 NAs.
The racial resentment outcomes are coded as binary, with "agree" as 1 and "disagree" as 0 for both questions below. Because of safeguard algorithms or biases in LLMs as discussed in the paper, they cannot replicate well questions related to racism:
Racial resentment 1: 
``Irish, Italian, Jewish and many other minorities overcame prejudice and worked their way up. Blacks should do the same without any special favors."
Do you agree strongly, agree somewhat, neither agree nor disagree, disagree somewhat, or disagree strongly with this statement?\\
Racial resentment 2: 
``Generations of slavery and discrimination have created conditions that make it difficult for blacks to work their way out of the lower class."
Do you agree strongly, agree somewhat, neither agree nor disagree, disagree somewhat, or disagree strongly with this statement?
The model's performance is notably less accurate for respondents who selected ``neither agree nor disagree." This discrepancy shows that demographic variables, party ID, and ideology (which generally serve as strong predictors) are insufficient to account for these neutral responses. A likely explanation is that respondents who select ``neither agree nor disagree" do so for heterogeneous and unobserved reasons, rather than due to an underlying ideological or demographic pattern. which can be explored by future research.

In the terminology of \cite{Rubin1976MissingData}, this type of missingness is best characterized as \textit{Missing Not at Random }(MNAR), which means that the decision to provide a neutral response is systematically related to factors beyond the observed covariates. This distinction underscores a key limitation in applying predictive models to survey responses: while strong ideological or demographic anchors enhance predictive accuracy for polarized positions, they may fail to capture the complexities underlying ambivalent or strategically neutral responses.
Moreover, these results are also comforting as they demonstrate the high level of transparency and scientific rigor achieved with large, representative datasets. This shows that CES and ANES data consistently measure the preferences of ordinary citizens, despite differences in sampling methods and research teams.

\section{Appendix D: Deep Learning Architecture}

I deliberately employ the most basic deep learning architecture to demonstrate that even a simple model: one that can be run in Google Colab via a URL in just a few seconds—can successfully transfer knowledge across survey datasets in American politics. In contrast, fine-tuning large language models is expensive, environmentally unsustainable, and often inaccessible to most researchers. This simple deep learning framework, when combined with established theories of political behavior, highlights how powerful yet accessible methods can already be. Future researchers can extend this foundation by employing more sophisticated neural architectures or by integrating imputation strategies to improve results. In the long run, this approach could also be packaged into easy-to-use software, enabling pollsters and scholars alike to upload their data and instantly generate silicon samples in a transparent and replicable way.

\begin{table}[h!]
\centering
\caption{Neural network architecture across three STL stages}
\scriptsize 
\begin{tabularx}{\textwidth}{lXXX}
\toprule
\textbf{Layer} & \textbf{Stage 1: CES Pretraining} & \textbf{Stage 2: ANES Fine-tuning} & \textbf{Stage 3: CES 2022 Silicon} \\
\midrule
Input       & 7 features (standardized + encoded) & Same (ANES features) & Same (CES 2022 features) \\
Dense 1     & 16 (ReLU), L2(0.01), BN, Dropout(0.3), trainable & Frozen (CES weights) & Frozen (reused backbone) \\
Dense 2     & 64 (ReLU), L2(0.01), BN, Dropout(0.4), trainable & Frozen (CES weights) & Frozen (reused backbone) \\
Dense 3     & 32 (ReLU), L2(0.01), BN, Dropout(0.4), trainable & Frozen (CES weights) & Frozen (reused backbone) \\
Dense 4     & 32 (ReLU), L2(0.01), BN, Dropout(0.3), trainable & Fine-tuned & Frozen (reused backbone) \\
Dense 5     & 16 (ReLU), L2(0.01), BN, Dropout(0.3), trainable & Fine-tuned & Frozen (reused backbone) \\
Dense 6     & 8 (ReLU),  L2(0.01), BN, Dropout(0.2), trainable & Fine-tuned & Frozen (reused backbone) \\
Output Head & 1 (Sigmoid), binary outcome & New head $h_2$ (Sigmoid), ANES outcome & New head $h_3$ (Sigmoid), synthetic outcome \\
\bottomrule
\end{tabularx}
\label{tab:stl_architecture}
\end{table}

\newpage





\newpage


\bibliographystyle{apalike}
\bibliography{sn-article}

\begin{thebibliography}{}

\bibitem[aap, 2025]{aapor2025}
 (2025).
\newblock Aapor 80th annual conference: Reshaping democracy’s oracle.
\newblock In {\em 80th Annual Conference of the American Association for Public Opinion Research (AAPOR)}.
\newblock Conference Chair: Gina Walejko; Associate Chair: Morgan Earp.

\bibitem[Abadi et~al., 2016]{abadi2016tensorflow}
Abadi, M. et~al. (2016).
\newblock Tensorflow: Large-scale machine learning on heterogeneous distributed systems.
\newblock {\em arXiv preprint arXiv:1603.04467}.

\bibitem[Abramowitz and Saunders, 2008]{Abramowitz2008}
Abramowitz, A.~I. and Saunders, K.~L. (2008).
\newblock Is polarization a myth?
\newblock {\em The Journal of Politics}, 70(2):542--555.

\bibitem[Agnew et~al., 2024]{agnew2024illusion}
Agnew, W., Bergman, A.~S., Chien, J., D{\'\i}az, M., El-Sayed, S., Pittman, J., Mohamed, S., and McKee, K.~R. (2024).
\newblock The illusion of artificial inclusion.
\newblock In {\em Proceedings of the CHI Conference on Human Factors in Computing Systems (CHI 2024)}.
\newblock Also available as arXiv preprint arXiv:2401.08572 [cs.CY].

\bibitem[Aher et~al., 2023]{aher2023using}
Aher, G.~V., Arriaga, R.~I., and Kalai, A.~T. (2023).
\newblock Using large language models to simulate multiple humans and replicate human subject studies.
\newblock In {\em Proceedings of the 40th International Conference on Machine Learning}, pages 337--371.

\bibitem[Argyle et~al., 2023]{argyle2023out}
Argyle, L.~P., Busby, E.~C., Fulda, N., Gubler, J.~R., Rytting, C., and Wingate, D. (2023).
\newblock Out of one, many: Using language models to simulate human samples.
\newblock {\em Political Analysis}, 31(3):337--351.

\bibitem[Bailey, 2024]{Bailey2024PollingAtACrossroads}
Bailey, M.~A. (2024).
\newblock {\em Polling at a Crossroads: Rethinking Modern Survey Research}.
\newblock Cambridge University Press.

\bibitem[Bartels, 2000]{bartels_2000}
Bartels, L.~M. (2000).
\newblock Partisanship and voting behavior, 1952–1996.
\newblock {\em American Journal of Political Science}, 44(1):35--50.

\bibitem[Benson, 2024]{Benson2024}
Benson, J. (2024).
\newblock Democracy and the epistemic problems of political polarization.
\newblock {\em American Political Science Review}, 118(4):1719--1732.

\bibitem[Berelson et~al., 1954]{berelson_lazarsfeld_mcphee_1954}
Berelson, B.~R., Lazarsfeld, P.~F., and McPhee, W.~N. (1954).
\newblock {\em Voting}.
\newblock University of Chicago Press, Chicago.

\bibitem[Bisbee et~al., 2024]{bisbee2024synthetic}
Bisbee, J., Clinton, J.~D., Dorff, C., Kenkel, B., and Larson, J.~M. (2024).
\newblock Synthetic replacements for human survey data? the perils of large language models.
\newblock {\em Political Analysis}, 32(4):401--416.

\bibitem[Bishop, 2008]{bishop_2008}
Bishop, B. (2008).
\newblock {\em The Big Sort: Why the Clustering of Like-Minded America Is Tearing Us Apart}.
\newblock Houghton Mifflin Harcourt, New York.

\bibitem[Broska et~al., 2025]{Broska2025}
Broska, D., Howes, M., and van Loon, A. (2025).
\newblock The mixed subjects design: Treating large language models as potentially informative observations.
\newblock \url{https://ssrn.com/abstract=5133034}.
\newblock MIT Sloan Research Paper No. 7154-24. \newline Available at SSRN: \url{https://ssrn.com/abstract=5133034} or \doi{10.2139/ssrn.5133034}.

\bibitem[Burns, 2006]{Burns2006History}
Burns, N. (2006).
\newblock Anes history.
\newblock \url{https://electionstudies.org/wp-content/uploads/2018/07/20060815Burns\_ANES\_history.pdf}.
\newblock Accessed: [Insert date here].

\bibitem[Campbell et~al., 1960]{campbell_converse_miller_stokes_1960}
Campbell, A., Converse, P.~E., Miller, W.~E., and Stokes, D.~E. (1960).
\newblock {\em The American Voter}.
\newblock John Wiley \& Sons, Inc., New York.

\bibitem[Chao et~al., 2023]{chao2023editorial}
Chao, G., Zhu, X., Ding, W., et~al. (2023).
\newblock Editorial: Special issue on transfer learning.
\newblock {\em Neural Processing Letters}, 55:1997--2000.

\bibitem[Chen et~al., 2020]{chen2020hybridqa}
Chen, W., Zha, H., Chen, Z., Xiong, W., Wang, H., and Wang, W.~Y. (2020).
\newblock Hybridqa: A dataset of multi-hop question answering over tabular and textual data.
\newblock In {\em Findings of the Association for Computational Linguistics: EMNLP 2020}, pages 1026--1036. Association for Computational Linguistics.

\bibitem[Clinton and Lapinski, 2024]{clinton2024polls}
Clinton, J. and Lapinski, J. (2024).
\newblock Once again, polls missed a decisive slice of trump voters in 2024.
\newblock Accessed: 2025-01-27.

\bibitem[Clinton et~al., 2021]{clinton2021aapor}
Clinton, J.~D. et~al. (2021).
\newblock American association of public opinion research task force on pre-election polling: An evaluation of the 2020 general election polls.
\newblock Technical report, American Association of Public Opinion Research.

\bibitem[Clinton et~al., 2022]{clinton2022reluctant}
Clinton, J.~D., Lapinski, J.~S., and Trussler, M.~J. (2022).
\newblock Reluctant republicans, eager democrats?: Partisan nonresponse and the accuracy of 2020 presidential pre-election telephone polls.
\newblock {\em Public Opinion Quarterly}, 86(2):247--269.

\bibitem[{Connected Papers}, 2025]{connectedpapers_not_asked_2025}
{Connected Papers} (2025).
\newblock Not asked and not answered: Multiple imputation for multiple surveys.
\newblock \url{https://www.connectedpapers.com/main/ec99899f958665e34d950856d2bf70d53ab3b204/Not-Asked-and-Not-Answered%3A-Multiple-Imputation-for-Multiple-Surveys/graph}.
\newblock Accessed: 2025-02-07.

\bibitem[Demszky et~al., 2023]{Demszky2023}
Demszky, D., Yang, D., Yeager, D.~S., Bryan, C.~J., Clapper, M., Chandhok, S., Eichstaedt, J.~C., Hecht, C., Jamieson, J., Johnson, M., Jones, M., Krettek-Cobb, D., Lai, L., Jones-Mitchell, N., Ong, D.~C., Dweck, C.~S., Gross, J.~J., and Pennebaker, J.~W. (2023).
\newblock Using large language models in psychology.
\newblock {\em Nature Reviews Psychology}, 2:688--701.

\bibitem[Fieck, 2025]{Fieck2025BiasInLLMs}
Fieck, S.~T. (2025).
\newblock An analysis of bias towards women in large language models using likert scale evaluations.
\newblock M.s. thesis, Chapman University, Orange, CA.

\bibitem[Fiorina et~al., 2011]{fiorina_abrams_pope_2011}
Fiorina, M.~P., Abrams, S.~J., and Pope, J.~C. (2011).
\newblock {\em Culture War?}
\newblock Longman.

\bibitem[Gelman et~al., 1999]{gelman1999notasked}
Gelman, A., King, G., and Liu, C. (1999).
\newblock Not asked and not answered: Multiple imputation for multiple surveys.
\newblock {\em Journal of the American Statistical Association}, 93(443):846--857.

\bibitem[Gimpel and Hui, 2015]{gimpel_hui_2015}
Gimpel, J. and Hui, I. (2015).
\newblock Seeking politically compatible neighbors? the role of neighborhood partisan composition in residential sorting.
\newblock {\em Political Geography}, 35(1):1--13.

\bibitem[Goodfellow et~al., 2016]{Goodfellow2016}
Goodfellow, I., Bengio, Y., and Courville, A. (2016).
\newblock {\em Deep Learning}.
\newblock The MIT Press.

\bibitem[Grossmann et~al., 2023]{grossmann2023ai}
Grossmann, I., Feinberg, M., Parker, D.~C., Christakis, N.~A., Tetlock, P.~E., and Cunningham, W.~A. (2023).
\newblock Ai and the transformation of social science research.
\newblock {\em Science}, 380(6650):1108--1109.

\bibitem[Hao et~al., 2024]{hao2024synthetic}
Hao, S., Han, W., Jiang, T., Li, Y., Wu, H., Zhong, C., Zhou, Z., and Tang, H. (2024).
\newblock Synthetic data in ai: Challenges, applications, and ethical implications.
\newblock {\em arXiv}, cs.LG.
\newblock arXiv:2401.01629v1.

\bibitem[Hetherington, 2009]{hetherington_2009}
Hetherington, M.~J. (2009).
\newblock Putting polarization in perspective.
\newblock {\em British Journal of Political Science}, 39(2):413--448.

\bibitem[Hillygus et~al., 2014]{hillygus2014professional}
Hillygus, D.~S., Jackson, N., and Young, M. (2014).
\newblock Professional respondents in nonprobability online panels.
\newblock In {\em Online Panel Research: Data Quality Perspective, A}, pages 219--237.

\bibitem[Hobbs and Vignoles, 2010]{Hobbs2010}
Hobbs, G. and Vignoles, A. (2010).
\newblock Is children’s free school meal ‘eligibility’ a good proxy for family income?
\newblock {\em British Educational Research Journal}, 36(4):673--690.
\newblock Accessed 17 Apr. 2025.

\bibitem[Iyengar et~al., 2019]{Iyengar2019}
Iyengar, S., Lelkes, Y., Levendusky, M., Malhotra, N., and Westwood, S.~J. (2019).
\newblock The origins and consequences of affective polarization in the united states.
\newblock {\em Annual Review of Political Science}, 22:129--146.

\bibitem[Josifoski et~al., 2023]{Josifoski2023}
Josifoski, M., Sakota, M., Peyrard, M., and West, R. (2023).
\newblock Exploiting asymmetry for synthetic training data generation: Synthie and the case of information extraction.
\newblock {\em arXiv preprint}, 2303.04132.

\bibitem[Keeter et~al., 2017]{keeter2017lowresponse}
Keeter, S., Hatley, N., Kennedy, C., and Lau, A. (2017).
\newblock What low response rates mean for telephone surveys.
\newblock Accessed: 2025-02-07.

\bibitem[Khalid et~al., 2024]{Khalid2024}
Khalid, M. et~al. (2024).
\newblock Novel sentiment majority voting classifier and transfer learning-based feature engineering for sentiment analysis of deepfake tweets.
\newblock {\em IEEE Access}, 12:67117--67129.

\bibitem[King et~al., 2001]{King2001Multiple}
King, G., Honaker, J., O'Connell, A.~J., and Scheve, K.~F. (2001).
\newblock Analyzing incomplete political science data: An alternative algorithm for multiple imputation.
\newblock {\em American Political Science Review}, 95(1):49--69.
\newblock Available at SSRN: \url{https://ssrn.com/abstract=1083698}.

\bibitem[King et~al., 2021]{king2021designing}
King, G., Keohane, R.~O., and Verba, S. (2021).
\newblock {\em Designing Social Inquiry: Scientific Inference in Qualitative Research, New Edition}.
\newblock Princeton University Press, Princeton, 2nd edition.
\newblock Publisher's Version Copy at \url{https://tinyurl.com/yd7gvcl2}.

\bibitem[Kreuter et~al., 2010]{Kreuter2010}
Kreuter, F., Olson, K.~M., Wagner, J., Yan, T., Ezzati-Rice, T.~M., Casas-Cordero, C., Lemay, M., Peytchev, A., Groves, R.~M., and Raghunathan, T.~E. (2010).
\newblock Using proxy measures and other correlates of survey outcomes to adjust for non-response: Examples from multiple surveys.
\newblock Sociology Department, Faculty Publications 137, University of Nebraska - Lincoln.

\bibitem[Laurer et~al., 2024]{Laurer2024}
Laurer, M., van Atteveldt, W., Casas, A., and Welbers, K. (2024).
\newblock Less annotating, more classifying: Addressing the data scarcity issue of supervised machine learning with deep transfer learning and bert-nli.
\newblock {\em Political Analysis}, 32(1):84--100.

\bibitem[Lazarsfeld et~al., 1944]{lazarsfeld_berelson_gaudet_1944}
Lazarsfeld, P.~F., Berelson, B.~R., and Gaudet, H. (1944).
\newblock {\em The People’s Choice}.
\newblock Duell, Sloan, and Pearce, New York.

\bibitem[Lenzerini, 2002]{10.1145/543613.543644}
Lenzerini, M. (2002).
\newblock Data integration: a theoretical perspective.
\newblock In {\em Proceedings of the Twenty-First ACM SIGMOD-SIGACT-SIGART Symposium on Principles of Database Systems}, PODS '02, page 233–246, New York, NY, USA. Association for Computing Machinery.

\bibitem[Levendusky, 2009]{levendusky_2009}
Levendusky, M. (2009).
\newblock {\em The Partisan Sort: How Liberals Became Democrats and Conservatives Became Republicans}.
\newblock University of Chicago Press.

\bibitem[Li et~al., 2023]{Li2023}
Li, Z., Zhu, H., Lu, Z., and Yin, M. (2023).
\newblock Synthetic data generation with large language models for text classification: Potential and limitations.
\newblock {\em arXiv preprint}, 2310.07849.
\newblock Presented at EMNLP 2023.

\bibitem[{Loughborough University Business School}, 2023]{lboro2023co2tool}
{Loughborough University Business School} (2023).
\newblock World first: Researchers create co\textsubscript{2} measurement tool to calculate emissions caused by digital data.
\newblock Accessed: 2025-02-07.

\bibitem[Ma and Safikhani, 2022]{ma2022theoretical}
Ma, M. and Safikhani, A. (2022).
\newblock Theoretical analysis of deep neural networks for temporally dependent observations.
\newblock In {\em Advances in Neural Information Processing Systems (NeurIPS)}.

\bibitem[Matthijsse et~al., 2015]{matthijsse2015internet}
Matthijsse, S.~M., De~Leeuw, E.~D., and Hox, J.~J. (2015).
\newblock Internet panels, professional respondents, and data quality.
\newblock {\em Methodology}.

\bibitem[McKenna et~al., 2023]{mckenna-etal-2023-sources}
McKenna, N., Li, T., Cheng, L., Hosseini, M., Johnson, M., and Steedman, M. (2023).
\newblock Sources of hallucination by large language models on inference tasks.
\newblock In Bouamor, H., Pino, J., and Bali, K., editors, {\em Findings of the Association for Computational Linguistics: EMNLP 2023}, pages 2758--2774, Singapore. Association for Computational Linguistics.

\bibitem[Meterko et~al., 2015]{meterko2015response}
Meterko, M., Restuccia, J.~D., Stolzmann, K., Mohr, D., Brennan, C., Glasgow, J., and Kaboli, P. (2015).
\newblock Response rates, nonresponse bias, and data quality: Results from a national survey of senior healthcare leaders.
\newblock {\em Public Opinion Quarterly}, 79(1):130--144.

\bibitem[Pan and Yang, 2010]{pan2010survey}
Pan, S.~J. and Yang, Q. (2010).
\newblock A survey on transfer learning.
\newblock {\em IEEE Transactions on Knowledge and Data Engineering}, 22(10):1345--1359.

\bibitem[Pilati et~al., 2024]{Pilati2024}
Pilati, F., Munk, A.~K., and Venturini, T. (2024).
\newblock The problems of llm-generated data in social science research.
\newblock {\em Sociologica}, 18(2):145--168.
\newblock Symposium: Repurposing Generative AI for Social Research – peer-reviewed.

\bibitem[Pokotylo, 2024]{pokotylo2024ethical}
Pokotylo, P. (2024).
\newblock Ethical and legal considerations of synthetic data usage.
\newblock \url{https://keymakr.com/blog/ethical-and-legal-considerations-of-synthetic-data-usage/}.
\newblock Keymakr Blog. Accessed: 2025-01-27.

\bibitem[Radford et~al., 2018]{radford2018gpt1}
Radford, A., Narasimhan, K., Salimans, T., and Sutskever, I. (2018).
\newblock Improving language understanding by generative pre-training.
\newblock Technical report, OpenAI.
\newblock \url{https://openai.com/research/language-unsupervised}.

\bibitem[Rapeli, 2022]{Rapeli2022}
Rapeli, L. (2022).
\newblock What is the best proxy for political knowledge in surveys?
\newblock {\em PLOS ONE}, 17(8):e0272530.

\bibitem[Ren et~al., 2024]{Ren2024LLMEnvironment}
Ren, S., Tomlinson, B., Black, R.~W., et~al. (2024).
\newblock Reconciling the contrasting narratives on the environmental impact of large language models.
\newblock {\em Scientific Reports}, 14:26310.

\bibitem[Rossi et~al., 2024]{Rossi2024}
Rossi, L., Harrison, K., and Shklovski, I. (2024).
\newblock The problems of llm-generated data in social science research.
\newblock {\em Sociologica}, 18(2):145--168.

\bibitem[Rubin, 1987]{Rubin1987Multiple}
Rubin, D. (1987).
\newblock {\em Multiple imputation for nonresponse in surveys}.
\newblock John Wiley and Sons, New York.

\bibitem[Rubin, 1976]{Rubin1976MissingData}
Rubin, D.~B. (1976).
\newblock Inference and missing data.
\newblock {\em Biometrika}, 63:581--592.

\bibitem[Rubin, 1996]{Rubin1996}
Rubin, D.~B. (1996).
\newblock Multiple imputation after 18+ years.
\newblock {\em Journal of the American Statistical Association}, 91(434):473--489.

\bibitem[Sarstedt et~al., 2024]{Sarstedt2024}
Sarstedt, M., Adler, S.~J., Rau, L., and Schmitt, B. (2024).
\newblock Using large language models to generate silicon samples in consumer and marketing research: Challenges, opportunities, and guidelines.
\newblock {\em Psychology \& Marketing}, 41(6):1254--1270.

\bibitem[Schafer, 1999]{Schafer1999}
Schafer, J.~L. (1999).
\newblock Multiple imputation: A primer.
\newblock {\em Statistical Methods in Medical Research}, 8(1):3--15.

\bibitem[Schmidt-Hieber, 2020a]{schmidt-hieber2020nonparametric}
Schmidt-Hieber, J. (2020a).
\newblock Nonparametric regression using deep neural networks with relu activation function.
\newblock {\em The Annals of Statistics}, 48(4):1875--1897.

\bibitem[Schmidt-Hieber, 2020b]{schmidt-hieber2020rejoinder}
Schmidt-Hieber, J. (2020b).
\newblock Rejoinder: ``nonparametric regression using deep neural networks with relu activation function''.
\newblock {\em The Annals of Statistics}, 48(4):1916--1921.

\bibitem[Shaha and Pawar, 2018]{Shaha2018}
Shaha, M. and Pawar, M. (2018).
\newblock Transfer learning for image classification.
\newblock In {\em 2018 Second International Conference on Electronics, Communication and Aerospace Technology (ICECA)}, pages 656--660.

\bibitem[Shapiro, 2019]{Shapiro2019}
Shapiro, W. (2019).
\newblock The polling industry is in crisis.
\newblock {\em New Republic}.

\bibitem[Simpson et~al., 2025]{Simpson2025ParityBenchmark}
Simpson, S., Nukpezah, J., Brooks, K., Pandya, R., et~al. (2025).
\newblock Parity benchmark for measuring bias in llms.
\newblock {\em AI and Ethics}, 5:3087--3101.

\bibitem[{Society for Political Methodology}, xlab]{polmeth_mailing_list}
{Society for Political Methodology} (\natexlab{}).
\newblock Polmeth mailing list.
\newblock \url{https://polmeth.org/mailing-list}.

\bibitem[{Stanford Institute for Research in the Social Sciences}, 2025]{anes_stanford}
{Stanford Institute for Research in the Social Sciences} (2025).
\newblock American national election studies.
\newblock Accessed: 2025-02-07.

\bibitem[Sui et~al., 2024]{sui2024table}
Sui, Y., Zhou, M., Zhou, M., Han, S., and Zhang, D. (2024).
\newblock Table meets llm: Can large language models understand structured table data? a benchmark and empirical study.
\newblock In {\em Proceedings of the 17th ACM International Conference on Web Search and Data Mining (WSDM '24)}, page~10, Mérida, Yucatán, Mexico. Association for Computing Machinery.

\bibitem[Sussell, 2013]{sussell_2013}
Sussell, J. (2013).
\newblock New support for the big sort hypothesis: An assessment of partisan geographic sorting in california, 1992--2010.
\newblock {\em PS: Political Science and Politics}, 46(4):768--773.

\bibitem[Taylor and Stone, 2009]{taylor2009transfer}
Taylor, M.~E. and Stone, P. (2009).
\newblock Transfer learning for reinforcement learning domains: A survey.
\newblock {\em Journal of Machine Learning Research}, 10:1633--1685.
\newblock Submitted 6/08; Revised 5/09; Published 7/09.

\bibitem[Torres and Cantú, 2022]{Torres2022}
Torres, M. and Cantú, F. (2022).
\newblock Learning to see: Convolutional neural networks for the analysis of social science data.
\newblock {\em Political Analysis}, 30(1):113--131.

\bibitem[Tripp and Dion, 2024]{tripp_dion_2024}
Tripp, A. and Dion, M. (2024).
\newblock American political science review: Annual editorial report.
\newblock \url{https://apsanet.org/wp-content/uploads/2024/10/APSR-2023-24-Annual-Report-.pdf}.
\newblock Prepared on behalf of the Editors, University of Wisconsin-Madison and McMaster University.

\bibitem[{United Nations Environment Programme}, 2024]{UNEP2024AIEnvironment}
{United Nations Environment Programme} (2024).
\newblock Ai has an environmental problem. here’s what the world can do about that.
\newblock \url{https://www.unep.org/news-and-stories/story/ai-has-environmental-problem-heres-what-world-can-do-about}.
\newblock Accessed: 2025-08-19.

\bibitem[van Buuren et~al., 2023]{mice}
van Buuren, S., Groothuis-Oudshoorn, K., Vink, G., Schouten, R., Robitzsch, A., Rockenschaub, P., Doove, L., Jolani, S., Moreno-Betancur, M., White, I., Gaffert, P., Meinfelder, F., Gray, B., Arel-Bundock, V., Cai, M., Volker, T., Costantini, E., van Lissa, C., and Oberman, H. (2023).
\newblock {\em Package 'mice'}.
\newblock R package version 3.16.0.

\bibitem[Veselovsky et~al., 2023]{Veselovsky2023}
Veselovsky, V., Ribeiro, M.~H., Arora, A., Josifoski, M., Anderson, A., and West, R. (2023).
\newblock Generating faithful synthetic data with large language models: A case study in computational social science.
\newblock {\em arXiv preprint}, 2305.15041.
\newblock 8 pages.

\bibitem[Walsh, 2024]{walsh2024tracking}
Walsh, C. (2024).
\newblock Tracking the american electorate: The cooperative election study.
\newblock \url{https://www.iq.harvard.edu/news/steve-ansolabehere-2024-election-feature}.
\newblock Institute for Quantitative Social Science, Harvard University. Accessed: 2025-02-07.

\bibitem[Weber and Klar, 2019]{weber_klar_2019}
Weber, C. and Klar, S. (2019).
\newblock Exploring the psychological foundations of ideological and social sorting.
\newblock {\em Political Psychology}, 40(S1):215--243.

\bibitem[Weidinger et~al., 2022]{weidinger2022taxonomy}
Weidinger, L., Uesato, J., Rauh, M., Griffin, C., Huang, P.-S., Mellor, J., Glaese, A., Cheng, M., Balle, B., Kasirzadeh, A., Biles, C., Brown, S., Kenton, Z., Hawkins, W., Stepleton, T., Birhane, A., Hendricks, L.~A., Rimell, L., Isaac, W., and Gabriel, I. (2022).
\newblock Taxonomy of risks posed by language models.
\newblock In {\em Proceedings of the 2022 ACM Conference on Fairness, Accountability, and Transparency}, pages 214--229.

\bibitem[Yosinski et~al., 2014]{yosinski2014transferable}
Yosinski, J., Clune, J., Bengio, Y., and Lipson, H. (2014).
\newblock How transferable are features in deep neural networks?
\newblock In {\em Advances in Neural Information Processing Systems (NIPS)}, volume~27.

\bibitem[Yuan and Rizoiu, 2025]{Yuan2025}
Yuan, L. and Rizoiu, M.-A. (2025).
\newblock Generalizing hate speech detection using multi-task learning: A case study of political public figures.
\newblock {\em Computer Speech \& Language}, 89:101690.

\bibitem[Zewe, 2025]{zewe2025generative}
Zewe, A. (2025).
\newblock Explained: Generative ai’s environmental impact.
\newblock {\em MIT News}.
\newblock Accessed: 2025-05-26.

\bibitem[Zhao et~al., 2024]{Zhao2024}
Zhao, Z., Alzubaidi, L., Zhang, J., Duan, Y., and Gu, Y. (2024).
\newblock A comparison review of transfer learning and self-supervised learning: Definitions, applications, advantages and limitations.
\newblock {\em Expert Systems with Applications}, 242:122807.

\bibitem[Škare et~al., 2024]{skare2024digitalization}
Škare, M., Gavurova, B., and Porada-Rochon, M. (2024).
\newblock Digitalization and carbon footprint: Building a path to a sustainable economic growth.
\newblock {\em Technological Forecasting and Social Change}, 199:123045.

\end{thebibliography}


\end{document}